# Semantic Graph Unification for Industrial Digital Threads: Bridging 11 Heterogeneous Manufacturing Systems Through Ontology-Driven Knowledge Graphs

Grama Chethan

Siemens Digital Industries Software

---

**ABSTRACT**

Modern manufacturing enterprises operate a constellation of heterogeneous information systems — enterprise resource planning (ERP), manufacturing execution systems (MES), product lifecycle management (PLM), SCADA, quality management systems (QMS), supply chain management (SCM), and more — each maintaining its own data model, schema, and API surface. The resulting data silos prevent holistic analysis, delay root-cause investigation, and obstruct the end-to-end traceability that Industry 4.0 demands. Integrating these systems via point-to-point connectors scales as $O(n^2)$ and accumulates brittle, undocumented dependencies that are costly to maintain.

This paper presents a complete, open framework for semantic graph unification of industrial digital threads. At its core is an ontology-driven RDF knowledge graph that unifies data from 11 simulated source entry points — covering nine distinct data domains (plant hierarchy, automation/control, product classification, alarm management, ERP planning, PLM engineering, MES execution, IoT monitoring, and supply chain) — through a five-stage ETL pipeline with automated cross-system entity resolution spanning 97 `owl:sameAs` identity categories. The purpose-built ontology encompasses 78 RDFS classes, 108 object properties, and 243 data properties, drawing on ISA-95, OPC UA, eCl@ss, the Asset Administration Shell (AAS), IEC 61131-3 [11], RAMI 4.0, and six additional standards. A golden-triangle validation scheme ensures referential consistency across the unified graph at load time.

On top of the unified knowledge base, an automated statistical discovery engine applies nine strategy categories — including cross-station correlation, alarm coverage analysis, ECN impact assessment, and CUSUM/EWMA process drift detection — to surface actionable manufacturing insights that span system boundaries. A comparative baseline study provides the paper's primary empirical contribution: blocking 24 cross-system MCP tools reduces recall from 1.00 to 0.31 (F1 from 1.00 to 0.48), demonstrating that 69% of discoverable signals require cross-system graph joins unavailable to single-source analysis. A leave-one-out ablation shows that six of nine strategies contribute unique, non-redundant signals. Internal verification against a 65-signal ground-truth manifest (16 positive, 49 null) confirms correct pipeline operation (F1 = 1.00, 95% Clopper-Pearson CI [0.79, 1.00]); because the manifest was constructed by the development team, this constitutes verification rather than independent validation. The graph is exposed to large language model (LLM) agents through 287 Model Context Protocol (MCP) tools functioning as a SPARQL-native semantic layer. Five industry-specific templates — aerospace, consumer packaged goods (CPG), pharmaceuticals, medical devices, and turbine blade manufacturing — demonstrate schema stability across discrete and process manufacturing verticals. The turbine blade configuration is the least mature of the five, sharing most of its structure with the aerospace template and serving primarily as a proof of extensibility rather than a fully differentiated industry vertical.

## 1. INTRODUCTION

A typical discrete manufacturing plant is a tower of Babel. The production floor communicates through OPC UA-compli-ant PLCs and SCADA historians that aggregate sensor readings at millisecond granularity. Above the shop floor, an MES such as Opcenter or FactoryTalk schedules work orders, tracks lot genealogy, and records non-conformances, using ISA-95 (IEC 62264) as its conceptual backbone. Engineering intent lives in a PLM system — Teamcenter, Windchill, or ENOVIA — where bills-of-material, engineering change orders, and revision histories accumulate over decades. Business planning runs through an ERP instance, commonly SAP S/4HANA, whose material master, vendor master, and financial postings use a schema entirely distinct from the MES. Products are described in supplier catalogues using eCl@ss IRDI identifiers that the ERP and PLM systems rarely consume natively. Quality data flows through a dedicated QMS, supply chain events through an SCM or logistics platform, and predictive maintenance signals through a condition monitoring system with its own time-series store. Each system has evolved over years under independent governance, often with overlapping but inconsistent definitions of the same real-world entities: a "part" in ERP is not quite the same object as a "component" in PLM or a "product" in eCl@ss. The consequence is fragmentation — information that, assembled, would tell a coherent story about quality, yield, cost, and compliance, instead remains trapped in silos that require laborious manual export-and-reconcile cycles to bridge.

Industry 4.0 articulates a compelling vision for overcoming this fragmentation. The Reference Architecture Model Industrie 4.0 (RAMI 4.0) proposes a three-dimensional coordinate system — life-cycle/value-stream axis, hierarchy-level axis, and IT-layer axis — as a common framework for positioning all manufacturing assets and their digital representations [21]. The Asset Administration Shell (AAS), standardized by the Industrial Digital Twin Association (IDTA), operationalizes RAMI 4.0 by defining a digital twin envelope [28] with typed submodels, enabling machine-readable exchange of engineering, operational, and maintenance data across organizational boundaries [3], and extending the digital twin paradigm from individual assets [14] to building- and city-scale deployments [15]. The promise is an end-to-end digital thread: a continuous, auditable information linkage from design intent through manufacturing execution, quality assurance, and field service, enabling traceability queries such as "which production lots were affected by the non-conforming raw material batch received on date X?" to be answered in seconds rather than days. In practice, however, realizing this vision requires more than agreeing on a reference architecture. It demands concrete mechanisms for ingesting heterogeneous source data, resolving cross-system entity identity, expressing semantic relationships in a queryable knowledge representation, and surfacing insights to human analysts and AI agents alike.

Existing integration approaches each capture part of the picture but leave critical gaps. Enterprise Service Bus (ESB) architectures excel at reliable message routing and protocol translation, but they operate at the message level, not the knowledge level; they do not capture the semantic relationships between entities in different systems, and accumulated point-to-point adapter mappings scale as $O(n^2)$ in the number of systems [26]. Data lake approaches aggregate raw data from multiple sources into a common storage tier but shed the semantic context that distinguishes a part number in ERP from the same identifier in PLM — a loss that renders cross-system joins unreliable without extensive manual curation. Canonical data model (CDM) efforts attempt to impose a single normalized schema but inevitably sacrifice domain richness; the ISA-95 hierarchy is far richer than any generic CDM can represent without domain-specific extensions that reintroduce the heterogeneity problem. Federated SPARQL query approaches presuppose that all source systems expose SPARQL endpoints with sufficiently aligned schemas — an assumption that fails in industrial settings where most systems expose proprietary REST or OPC UA APIs, not SPARQL endpoints. Commercial integration platforms (iPaaS) offer low-code mapping tools but generate opaque, non-queryable pipelines and do not produce first-class knowledge graphs amenable to reasoning and graph analytics. None of these approaches simultaneously addresses semantic richness, automated cross-system entity resolution, and AI-ready access through a standardized tool interface — though each addresses a subset of these requirements in isolation.

This paper presents a unified framework that addresses these gaps through three mutually reinforcing contributions. First, we introduce a purpose-built ontology that spans 10+ industrial standards within a single coherent namespace, providing 78 RDFS classes and 351 properties as a semantic foundation for the digital thread. Second, we describe a five-stage ETL pipeline that ingests data from 11 simulated industrial systems — each exposing a different API — resolves cross-system entity identity through 97 `owl:sameAs` triples across seven boundary categories, and validates the resulting graph through a golden-triangle consistency check before persistence to an RDF triple store. Third, we present an automated discovery engine that applies nine statistical discovery strategies — with an architectural Bonferroni correction safeguard — to surface actionable insights from the unified graph, together with a Model Context Protocol (MCP) layer that exposes 287 SPARQL tools (264 hand-authored, 23 LLM-generated with schema validation) to LLM-based agents. The entire framework is validated across five industry-specific templates spanning discrete and process manufacturing.

Empirically, the pipeline produces a unified graph of approximately 27,800 RDF triples — comprising ~2,200 TBox schema triples (ontology definitions) and ~25,600 ABox instance triples generated by 11 simulated source entry points (some sharing underlying protocols — e.g., TIA Portal and OPC UA both model PLC-level automation; ISA-95 and Opcenter both model production execution — yielding nine distinct data domains: plant hierarchy, automation and control, product classification, alarm management, ERP planning, PLM engineering, MES execution, IoT monitoring, and supply chain management) — in approximately 5.2 seconds on commodity hardware. A comparative baseline study provides the paper's strongest result: blocking 24 cross-system MCP tools reduces recall from 1.00 to 0.31 (F1 from 1.00 to 0.48), demonstrating that 69% of discoverable signals require cross-system graph joins unavailable to single-source analysis. An internal verification run against a 65-signal ground-truth manifest confirms correct pipeline operation (F1 = 1.00, 95% Clopper-Pearson CI [0.79, 1.00]), though this constitutes verification rather than independent validation. The MCP tool layer supports natural language querying of the knowledge graph through 287 SPARQL-native tools functioning as a semantic layer. These results demonstrate that semantic unification at the scale of a manufacturing digital thread is both technically feasible and practically useful, narrowing the gap between the RAMI 4.0 vision and operational practice in heterogeneous manufacturing environments.

The remainder of the paper is organized as follows. Section 2 surveys related work in industrial ontologies, knowledge graph construction, cross-system integration, and AI-augmented manufacturing analytics. Section 3 presents the system architecture. Section 4 describes the ontology design, covering its class hierarchy, property axioms, design methodology, and alignment to external standards. Section 5 presents the five-stage ETL pipeline and entity resolution mechanisms. Section 6 details the automated discovery engine, its nine statistical strategies, and the multiple-comparison correction scheme. Section 7 describes multi-industry generalization across five manufacturing verticals. Section 8 reports evaluation results. Section 9 discusses limitations, and Section 10 concludes. The primary contributions of this work are:

1. An extensible ontology spanning 10+ industrial standards with 78 RDFS classes, 108 object properties, and 243 data properties, paired with a five-stage ETL pipeline that resolves cross-system entity identity through 97 `owl:sameAs` triples across seven boundary categories and golden-triangle validation (Sections 4–5).
2. Empirical evidence that cross-system graph unification is a prerequisite for the majority of discoverable manufacturing signals: a comparative baseline study shows that blocking cross-system joins reduces recall from 1.00 to 0.31 (F1 from 1.00 to 0.48), and a leave-one-out ablation demonstrates that six of nine strategies contribute unique, non-redundant findings (Section 8.7).
3. An automated discovery engine with nine statistical strategy categories — including cross-station correlation, alarm coverage analysis, ECN impact assessment, and CUSUM/EWMA process drift detection — verified against a 65-signal synthetic manifest (F1 = 1.00, 95% Clopper-Pearson CI [0.79, 1.00]; Section 6).
4. A 287-tool MCP layer functioning as a SPARQL-native semantic layer for LLM agents, with multi-industry generalization demonstrated across five manufacturing verticals through industry-specific ontology templates (Sections 6.6, 7).

## 2. RELATED WORK

### 2.1 Industrial Ontologies and Standards

The landscape of industrial standards provides rich domain vocabularies but remains fragmented across vertical silos. ISA-95 (IEC 62264) defines an enterprise-control integration model covering equipment hierarchy, production scheduling, personnel, and material information, and has been widely adopted as a conceptual backbone for MES interoperability [1]. Our ontology draws primarily on ISA-95 Part 2 (object model attributes for operations) and selected Part 4 elements (work order and material definitions); it does not implement the full Part 3 activity model or Part 5 B2MML transaction schemas. Mellor et al. [18] demonstrated an OWL mapping of ISA-95 Part 2 that enables SPARQL-based queries over production operations data, and Vegetti et al. [17] extended this to cover supply chain and product structure relationships. While these mappings faithfully represent ISA-95 semantics, they do not address the integration challenge of aligning ISA-95 entities with identically named but differently modeled concepts in PLM or ERP systems. OPC UA (IEC 62541) provides an information model for industrial automation that has become the de facto standard for device-level interoperability [2]. Its companion specifications — covering domains such as robotics, CNC machinery, and batch processing — extend the base model with domain-specific node types, and recent work has explored RDF serializations of OPC UA information models to enable SPARQL querying of automation data [22]. However, OPC UA's graph model is designed for real-time device communication, not for persistent semantic integration with enterprise systems. Notably, OPC 30530 (the ISA-95 Companion Specification for OPC UA) defines a standardized OPC UA information model that maps ISA-95 equipment, personnel, and material concepts to OPC UA node types [22], enabling device-level access to production-execution data. Our approach differs from OPC 30530 in three respects: (i) we unify non-OPC-UA systems — including SAP ERP, Teamcenter PLM, and supply-chain management — that fall outside OPC UA's automation scope; (ii) we provide explicit `owl:sameAs` entity resolution across system boundaries, whereas OPC 30530 relies on OPC UA's built-in type system without cross-system identity alignment; and (iii) the resulting RDF graph supports SPARQL-based reasoning and multi-hop traversal that OPC UA's hierarchical address space does not natively enable. OPC 30530 is thus complementary: it could serve as a more standardized source adapter for the ISA-95 and OED layers of our framework.

The Asset Administration Shell (AAS), standardized by IDTA, represents the most ambitious current effort toward a unified digital twin envelope for Industry 4.0 [3]. AAS defines a hierarchical structure of asset and submodel descriptions with typed properties, references, and operations, and has been extended through community- developed submodel templates covering areas such as technical data, carbon footprint, and simulation. Nevertheless, AAS is primarily a serialization format and an information exchange protocol; it does not provide the reasoner-accessible ontological axioms needed for cross-system entity resolution, nor does it specify mechanisms for integrating AAS instances with ERP or SCADA data outside the shell boundary. Bader et al. [22, 33] explored RDF serializations of AAS (AASX-to-RDF) and demonstrated SPARQL querying of AAS submodel data; our framework builds on this direction by embedding AAS-derived triples within a broader 11-source knowledge graph that contextualizes AAS digital twins with ERP, MES, and supply chain data. eCl@ss, with its hierarchical product classification and IRDI-based property definitions, provides a controlled vocabulary for product description that bridges procurement and engineering, but its uptake as a semantic layer in knowledge graphs remains limited [4]. RAMI 4.0 [21] serves as an architectural reference rather than an implementation specification; it positions the AAS within a three-dimensional coordinate system but leaves the semantic details of cross-standard alignment unspecified. The collective gap across all these standards is the absence of a unified, implementation-ready ontology that expresses cross-standard alignment axioms, supports automated entity resolution, and can be queried by both human analysts and AI agents through a consistent interface.

The **Industrial Ontology Foundry (IOF)** represents the most significant community effort to build a foundational ontology for manufacturing, grounded in the Basic Formal Ontology (BFO) upper ontology and aligned with the Common Core

Ontologies (CCO). IOF provides carefully axiomatized definitions of manufacturing concepts such as processes, artifacts, and capabilities, and has produced domain-specific modules for maintenance, quality, and assembly. Our work does not extend IOF for pragmatic reasons: (1) IOF's BFO grounding requires OWL DL reasoning, while our framework targets SPARQL-centric query workloads over an rdflib triple store that does not support OWL DL entailment; (2) IOF's current modules cover a subset of our ten-standard scope (ISA-95, OPC UA, ISA-18.2, AAS, eCl@ss, SAP, Teamcenter, Insights Hub, SCM, TIA Portal), and extending IOF to cover all ten would require community consensus timelines incompatible with our engineering objectives; and (3) our 78-class RDFS ontology is deliberately lightweight to support sub-second queries at the cost of formal expressivity. Alignment between our ontology and IOF modules remains a valuable future direction — our RDFS classes map straightforwardly to IOF counterparts where they exist, and a formal alignment would enable interoperability with the broader IOF ecosystem.

Three additional ontological efforts are relevant. **NGSI-LD** [41] (ETSI ISG CIM) provides an RDF-compatible information model for context information management, widely adopted in smart city and IoT deployments. NGSI-LD's entity-property-relationship model maps naturally to RDF triples, and its temporal property support is relevant to our time-series aspects; however, NGSI-LD focuses on IoT context brokering and does not address enterprise-level manufacturing concepts (BOM, routing, work orders, NCRs). The **MASON ontology** (Manufacturing's Semantics Ontology) [39] provides an OWL-based vocabulary for manufacturing processes, resources, and capabilities, grounded in DOLCE and covering concepts comparable to a subset of ISA-95. Our framework's scope extends well beyond MASON's manufacturing-process focus to include ERP planning, supply chain, product classification, and alarm management. **OntoSTEP** [40], developed at NIST, maps ISO 10303 (STEP) product data to OWL, providing formal semantics for CAD geometry, assembly structure, and product manufacturing information (PMI). OntoSTEP addresses product-data semantics that our framework currently handles through Teamcenter item revisions; a formal alignment between our `ont:ItemRevision` class and OntoSTEP's product model would strengthen the framework's connection to the broader STEP ecosystem.

**Microsoft's Digital Twins Definition Language (DTDL)**, used by Azure Digital Twins, provides a JSON-LD-based modeling language for describing digital twin instances and their relationships. DTDL v3 added semantic annotations (`@context` extensions) that allow models to reference external ontologies, partially addressing earlier criticism of semantic isolation. Nevertheless, DTDL remains architecturally different from RDF/RDFS: it lacks SPARQL query support, does not provide `owl:sameAs`-style entity resolution across heterogeneous sources, and is optimized for Azure-managed infrastructure rather than open, portable deployments. DTDL's expressivity is comparable to a property graph with typed edges — suitable for IoT device modeling but insufficient for the multi-standard ontological alignment required by our 11-source integration scope. The **W3C Web of Things (WoT) Thing Description (TD)** standard provides a JSON-LD vocabulary for describing IoT device capabilities (properties, actions, events) and protocol bindings. WoT TD is complementary to our approach: our IoT adapter (Insights Hub) could consume WoT TDs as a source format, and WoT's protocol binding metadata could enrich our OPC UA tag descriptions. However, WoT TD operates at the device interaction level and does not address the enterprise-level integration of ERP, PLM, MES, and supply chain systems that constitutes the core scope of our framework.

### 2.2 Knowledge Graph Construction

Knowledge graph construction from heterogeneous sources has been extensively surveyed by Hogan et al. [7], with complementary treatments by Sheth et al. [24] on knowledge graphs at scale and Biffl et al. [23] on semantic web technologies for engineering applications. These works provide a taxonomy covering knowledge representation, construction methodologies (extraction, curation, synthesis), and enrichment techniques including link prediction and ontology alignment. A central challenge in multi-source knowledge graph construction is entity resolution — the task of determining when records from different sources refer to the same real-world entity. Christophides et al. [8] survey entity resolution in knowledge graphs, distinguishing blocking strategies that reduce comparison space from matching strategies that assess similar-ity, and highlight the particular difficulty of resolving entities across schema heterogeneity. Schema alignment and ontology matching, surveyed by Euzenat and Shvaiko [9], provide formal frameworks for establishing correspondences between heterogeneous ontologies, but automated matching tools achieve limited precision in specialized industrial domains where standard string similarity metrics fail to capture domain semantics — for example, distinguishing an ERP "material" from a PLM "part" requires understanding organizational workflows, not just string patterns. Our work addresses entity resolution through a hybrid approach: a lexicon of 18 `owl:sameAs` link categories defined by domain experts and applied automatically by the ETL pipeline, with golden-triangle validation to catch resolution errors at load time rather than propagating them into the query layer.

Industrial knowledge graph initiatives from major automation vendors have demonstrated the value of semantic integration at scale, particularly in IIoT contexts [25]. Siemens MindSphere employed graph-based asset modeling for IoT data management, while Bosch IoT Suite has explored ontology-driven device management using W3C Thing Description. ABB Ability leverages knowledge graphs for cross-plant benchmarking of operational performance. Siemens' Industrial Knowledge Graph (IKG) [31] demonstrated large-scale RDF knowledge graph construction from heterogeneous industrial sources, achieving cross-domain linking of product, process, and resource data; our work extends this direction by adding automated statistical discovery and an MCP-based LLM query interface. The Blue Brain Project's **Knowledge Graph Forge** [32] provides a Python framework for building, managing, and querying knowledge graphs with typed resources and SHACL-based validation — addressing the general-purpose graph lifecycle problem. Our work specializes this pattern for the manufacturing domain with industry-specific ontology templates, cross-system entity resolution via `owl:sameAs`, and a discovery engine that operates over the resulting unified graph. Lécué [38] argues that knowledge graphs provide a natural substrate for explainable AI in industrial settings, offering structured provenance and interpretability that opaque ML models lack — a perspective directly relevant to our discovery engine's use of graph-derived evidence chains. These industrial deployments and research frameworks share a common gap: they do not provide open, extensible frameworks that combine semantic unification with automated cross-system discovery and LLM-native query interfaces. Our framework is explicitly designed for heterogeneous system integration using open standards, and it exposes the resulting knowledge base through a standardized AI tool interface.

Beyond vendor-specific platforms, several commercial integration suites offer manufacturing knowledge management capabilities. **PTC ThingWorx** provides a property-graph-based data model with REST APIs for IoT and manufacturing data, but its integration scope is typically limited to PTC's own PLM (Windchill) and IoT products and does not support SPARQL or formal ontological alignment across vendor boundaries. **Rockwell Automation FactoryTalk** focuses on OT-layer integration (Allen-Bradley PLCs, HMIs, historians) with limited IT-layer reach into ERP or PLM systems. **Dassault Systèmes 3DEXPERIENCE** provides a comprehensive PLM-centric platform with ENOVIA for configuration management and DELMIA for manufacturing operations, but its data model is proprietary and closed to external RDF-based querying. These commercial platforms share two structural limitations relative to our approach: (1) they are optimized for their own vendor ecosystem and impose significant integration effort for third-party systems, and (2) they do not expose their data through open semantic standards (RDF, SPARQL, OWL) that would enable vendor-neutral tooling and LLM-based query interfaces.

**Table 1. Qualitative comparison of industrial knowledge graph approaches. ✔ = supported, ○ = partial, ✘ = not supported.**

| Capability | This work | Siemens IKG [31] | KG Forge [32] | ThingWorx | 3DEXPERIENCE |
|---|---|---|---|---|---|
| Standards integrated | 10 | 3–5 | Generic | 1–2 | 2–3 |
| Open query (SPARQL) | ✔ | ✔ | ✔ | ✘ | ✘ |
| Cross-system entity resolution | ✔ | ○ | ○ | ✘ | ✘ |
| Automated statistical discovery | ✔ | ✘ | ✘ | ✘ | ✘ |
| LLM tool interface (MCP) | ✔ | ✘ | ✘ | ✘ | ✘ |
| Multi-industry templates | ✔ (5) | ○ | ✔ | ○ | ○ |
| Open-source / vendor-neutral | ✔ | ✘ | ✔ | ✘ | ✘ |

### 2.3 Cross-System Integration Approaches

The Enterprise Service Bus (ESB) pattern, epitomized by products such as MuleSoft Anypoint and IBM Integration Bus, provides reliable message routing, protocol translation, and transformation between enterprise systems [26]. ESBs have been widely deployed in manufacturing IT landscapes to connect ERP, MES, and PLM systems, but they operate at the syntactic level of message passing rather than the semantic level of knowledge representation. The adapter landscape for an ESB-integrated plant scales as $O(n^2)$ in the number of systems: adding the nth system requires up to $n-1$ new adapters, and each adapter encodes implicit assumptions about the source and target schemas that are difficult to audit or reuse. Canonical data model (CDM) approaches attempt to reduce this complexity by routing all transformations through a common schema, but any fixed CDM must choose between completeness (which leads to a schema too rich for any single domain to navigate) and focus (which loses domain-specific richness). Federated query approaches, including federated SPARQL over distributed endpoints [6], offer an alternative by leaving data in place and pushing query decomposition to a mediator. In manufacturing settings, however, most operational systems expose proprietary REST, OPC UA, or SOAP APIs rather than SPARQL endpoints, making federation impractical without per-system wrappers that introduce the very $O(n^2)$ adapter problem that federation was intended to solve.

Digital thread frameworks — conceptual architectures for end-to-end traceability from design through production to sustainment — have received growing attention in the aerospace and defense sector [13]. DARPA's Open Manufacturing program and subsequent efforts at NIST have produced reference implementations for specific product families, but comprehensive, open implementations that span the full system landscape of a discrete manufacturer remain rare. Most published digital thread demonstrations focus on a subset of the integration problem — typically the design-to-manufacturing interface — and do not address the integration of quality, supply chain, and maintenance data that a holistic digital thread requires. Our framework contributes a comparatively broad open implementation of a manufacturing digital thread, spanning 11 simulated systems across the full ISA-95 hierarchy from enterprise planning to device-level automation.

### 2.4 AI-Augmented Manufacturing Analytics

Statistical process control (SPC) and Six Sigma traditions have long provided rigorous methods for identifying assignable causes of variation in manufacturing processes, including Shewhart control charts, CUSUM analysis, and design-of- exper-iments (DoE) methods [16]. These methods, however, are designed for single- system, well-structured datasets, not for cross-system discovery over heterogeneous knowledge graphs. Machine learning has been applied to predictive maintenance [20] and quality prediction [20] in manufacturing with demonstrated industrial impact, but these applications typically operate within a single system's data silo and do not benefit from the cross-system context that a unified knowledge graph provides. Recent work on LLM-based manufacturing agents has explored using large language models to interpret process data and to generate executable process plans, but these efforts rely on unstructured or lightly structured data rather than on a semantically grounded knowledge base that constrains the agent's reasoning to validated facts.

A growing body of work addresses LLM-based SPARQL generation — the task of translating natural language questions into executable SPARQL queries. Benchmarks such as LC-QuAD 2.0 and DAIL-SQL evaluate this capability, and recent approaches including Text2SPARQL and schema-prompted code generation have achieved competitive results on open-domain knowledge graphs (e.g., Wikidata, DBpedia). However, these approaches face two challenges in industrial settings: (1) industrial ontologies are narrow-domain, proprietary, and absent from LLM training corpora, meaning the model has no prior exposure to the schema it must query; and (2) industrial queries frequently require multi-hop joins across 3–7 systems, with correctness depending on precise alignment edges ( `owl:sameAs` ) that raw SPARQL generation is unlikely to discover without explicit schema guidance. Our MCP tool-based approach avoids the SPARQL generation problem entirely: each tool encapsulates a validated, tested SPARQL query with typed inputs and structured outputs, reducing the LLM's task from query generation to tool selection — which we hypothesize to be a substantially easier planning problem, though this claim requires empirical validation comparing tool-selection accuracy against Text-to-SPARQL generation accuracy on the same query workload. The **GraphRAG** paradigm (Microsoft, 2024) extends retrieval-augmented generation to graph-structured knowledge by constructing community summaries over entity clusters and using them as retrieval units. While GraphRAG addresses the retrieval bottleneck for LLM reasoning over graphs, it operates over extracted entity-relation triples and does not preserve the formal ontological structure (class hierarchy, cardinality constraints, cross-system alignment axioms) that our framework maintains. Our approach is complementary: GraphRAG-style summarization could be layered atop our unified graph for exploratory question answering, while our structured tool interface serves the precision queries that manufacturing decision-making requires.

**Semantic layers** — including dbt metrics, Cube.js, and Looker LookML — provide a related pattern: curated, named metric definitions that abstract raw SQL behind typed interfaces, enabling BI tools and, increasingly, LLM agents to query data without generating SQL directly. Our MCP tool approach shares this architectural intuition (pre-validated queries with typed inputs), but differs in three respects: (1) semantic layers operate over relational schemas, while our tools operate over an RDF graph with formal ontological grounding (class hierarchy, domain/range inference, cross-system alignment via `owl:sameAs` ); (2) semantic layers expose aggregation metrics (revenue, churn rate), while our tools expose multi-hop graph traversals across 3–7 systems that have no natural relational equivalent; and (3) semantic layers assume a single data warehouse, while our framework unifies 11 heterogeneous sources with distinct schemas, APIs, and identity spaces. The tool layer could be viewed as a SPARQL-native semantic layer purpose-built for manufacturing knowledge graphs.

**LLM tool-use frameworks** provide complementary infrastructure for exposing structured tools to language model agents. LangChain's StructuredTool abstraction [34] wraps arbitrary functions with Pydantic schemas for input validation, while LlamaIndex provides query engine tools that mediate between natural language and structured data sources. Microsoft's Semantic Kernel [35] offers a similar pattern through its plugin architecture, and ToolBench [36] and API-Bank [37] provide benchmarks for evaluating LLM tool selection at scale (16,000+ and 53 tools, respectively). Our MCP tool layer shares the core architectural intuition of these frameworks — typed interfaces with schema-validated inputs — but differs in two key respects: (1) each tool encapsulates a pre-validated SPARQL query against a formal ontology, rather than wrapping arbitrary API endpoints; and (2) the tool set is specifically designed for multi-hop graph traversals across heterogeneous industrial systems, a use case not addressed by existing LLM tool benchmarks. The ToolBench finding that LLMs degrade in tool selection accuracy beyond ~50 simultaneously presented tools [36] is directly relevant to our 287-tool inventory and is discussed in Section 9.

The Model Context Protocol (MCP), introduced by Anthropic in 2024, provides a standardized interface for exposing tools and resources to LLM-based agents, enabling structured tool use that goes beyond raw text prompting [12]. MCP has been applied in software development and data analytics contexts, but its application to industrial knowledge graph access is novel. The combination of a semantically grounded knowledge graph with an MCP tool layer creates a system in which an LLM agent can issue well-formed SPARQL queries — mediated through curated and validated tool definitions — over a validated, cross-system knowledge base, grounding agent reasoning in auditable facts rather than in training-time statistical regularities. Our work extends this paradigm with an LLM-assisted tool authoring loop: when an LLM agent encounters a query pattern not covered by existing tools, the system generates a candidate SPARQL tool definition, validates it against the ontology schema, and persists successful definitions for reuse — incrementally expanding the agent's capability with human-reviewable artifacts. Over a five-month development period, this mechanism produced 23 validated tools (an 8% expansion of the hand-authored base) at a rate of approximately one tool per week; success and failure rates for the generation loop are not yet systematically tracked, which limits claims about the mechanism's reliability. No prior work, to our knowledge, combines semantic industrial knowledge graph unification with statistical multi-strategy discovery and an MCP-based semantic layer for LLM agents within a single open framework.

## 3. SYSTEM ARCHITECTURE

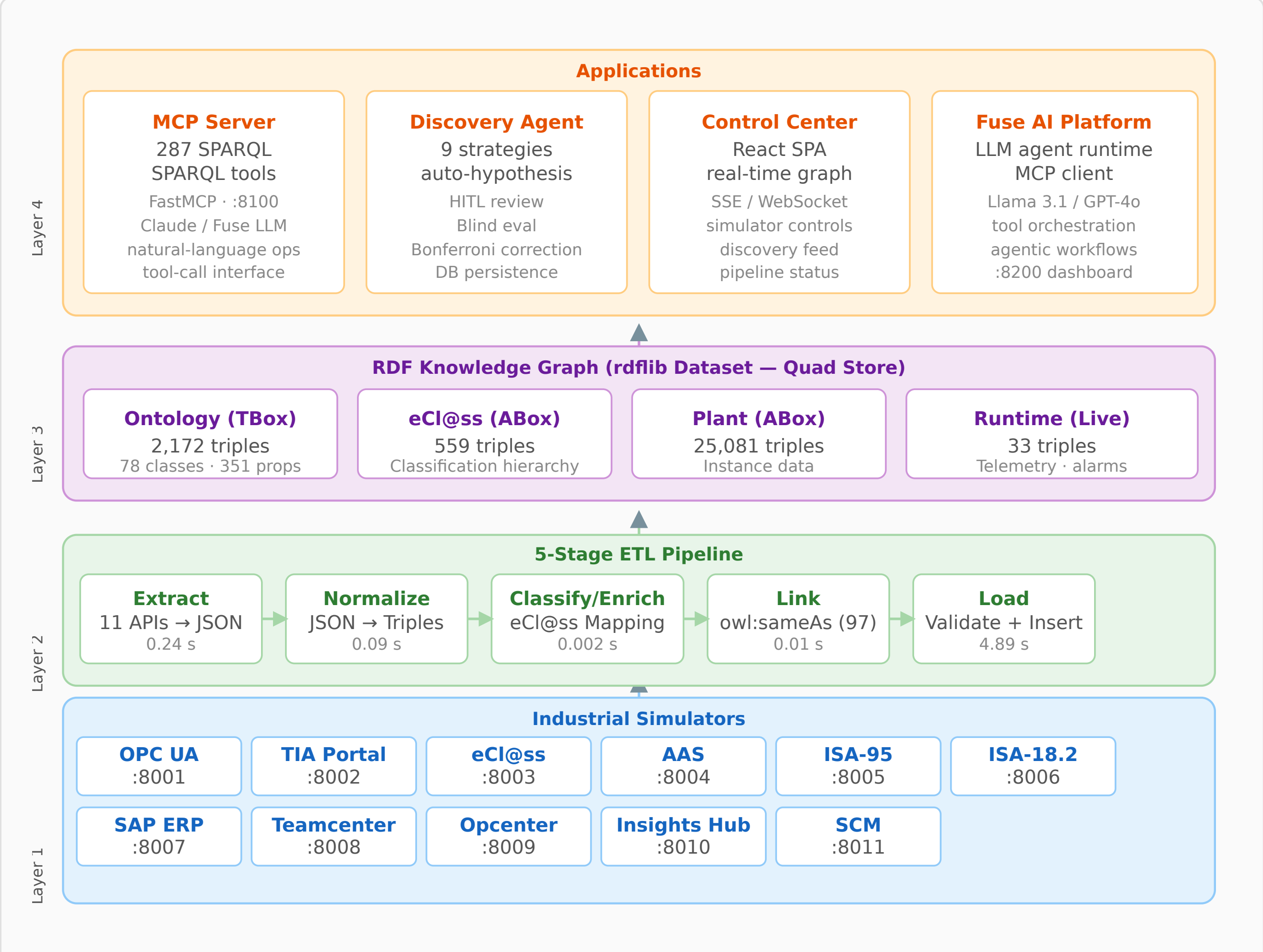

***Figure 1.*** *GraphUnifier four-layer architecture. Layer 1 (blue) consists of eleven independent industrial simulator REST services. Layer 2 (green) implements the five-stage ETL pipeline. Layer 3 (purple) is the rdflib quad-store partitioned into four named graphs. Layer 4 (orange) exposes the unified graph to downstream applications and AI agents.*

The system is organised as a four-layer stack (Figure 1) in which each layer communicates only with its immediate neighbours. The *GraphUnifier* coordinator owns the source-endpoint registry and orchestrates the five-stage ETL pipeline (Section 5), maintaining cumulative run statistics (triple counts, link counts, timing breakdowns). The *GraphStore* wraps an rdflib `Dataset` quad store partitioned into four named graphs by epistemic role: TBox axioms (ontology), eCl@ss classification hierarchy, ABox instance data from all eleven simulators (plant), and a rolling window of live telemetry (runtime). Thread-safe access via a reentrant lock ensures concurrent ETL cycles do not corrupt in-progress SPARQL queries.

The *SPARQLEngine* provides cross-graph federated query execution using SPARQL 1.1 `GRAPH` keywords, enabling queries that join across named graphs in a single round trip — e.g., correlating a live alarm with its equipment hierarchy and OPC UA tag type. The *Pipeline* executes stages sequentially with per-stage timing, publishes progress via an EventBus (SSE, WebSocket, and webhook bridges for real-time UI updates), and isolates partial failures so a simulator timeout does not corrupt graph state.

Layer 1 consists of eleven independent Python Flask services generating realistic synthetic industrial data, covering ISA-95 work orders, OPC UA tags and alarms, SAP production orders and routings, Teamcenter item revisions and BOMs, Insights Hub time-series readings, and SCM supplier records. A configurable *SimulationClock* provides time dilation (validated up to 60×) to ensure timestamp consistency across simulators during accelerated testing — a prerequisite for the discovery engine's temporal correlation strategies.

Table 1b maps the eleven source simulators to the seven industrial domains they collectively represent. Several domains are served by multiple simulators: for example, *Production Operations* spans ISA-95 (plant hierarchy and equipment), OED (work orders and operations), and SAP (production orders and material reservations). Conversely, the *Automation & Control* domain is split across OPC UA (tags and process variables), TIA Portal (PLC programs, HMI screens, PROFINET topology), and ISA-18.2 (alarm definitions and lifecycle). This many-to-many relationship between simulators and domains motivates the `owl:sameAs` linking strategy described in Section 5.

**Table 1b. Mapping of eleven source simulators to seven industrial domains.**

| Industrial Domain | Source Simulators | Key Entities |
|---|---|---|
| Production Operations | ISA-95, OED, SAP | WorkUnit, WorkOrder, Operation, ProductionOrder |
| Automation & Control | OPC UA, TIA Portal | Tag, PLCStation, HMIScreen, ProgramBlock |
| Alarm Management | ISA-18.2 | AlarmDefinition, AlarmPriority, AlarmKPI |
| Digital Twin | AAS | AASShell, Submodel, ConceptDescription |
| Product Classification | eCl@ss | EClassNode, classifiedAs relationships |
| IoT Monitoring | Insights Hub | IoTAsset, AspectType, TimeSeriesVariable, IoTEvent |
| Supply Chain / ERP / PLM | SAP, Teamcenter, SCM | Material, ItemRevision, Supplier, PurchaseOrder |

## 4. ONTOLOGY DESIGN

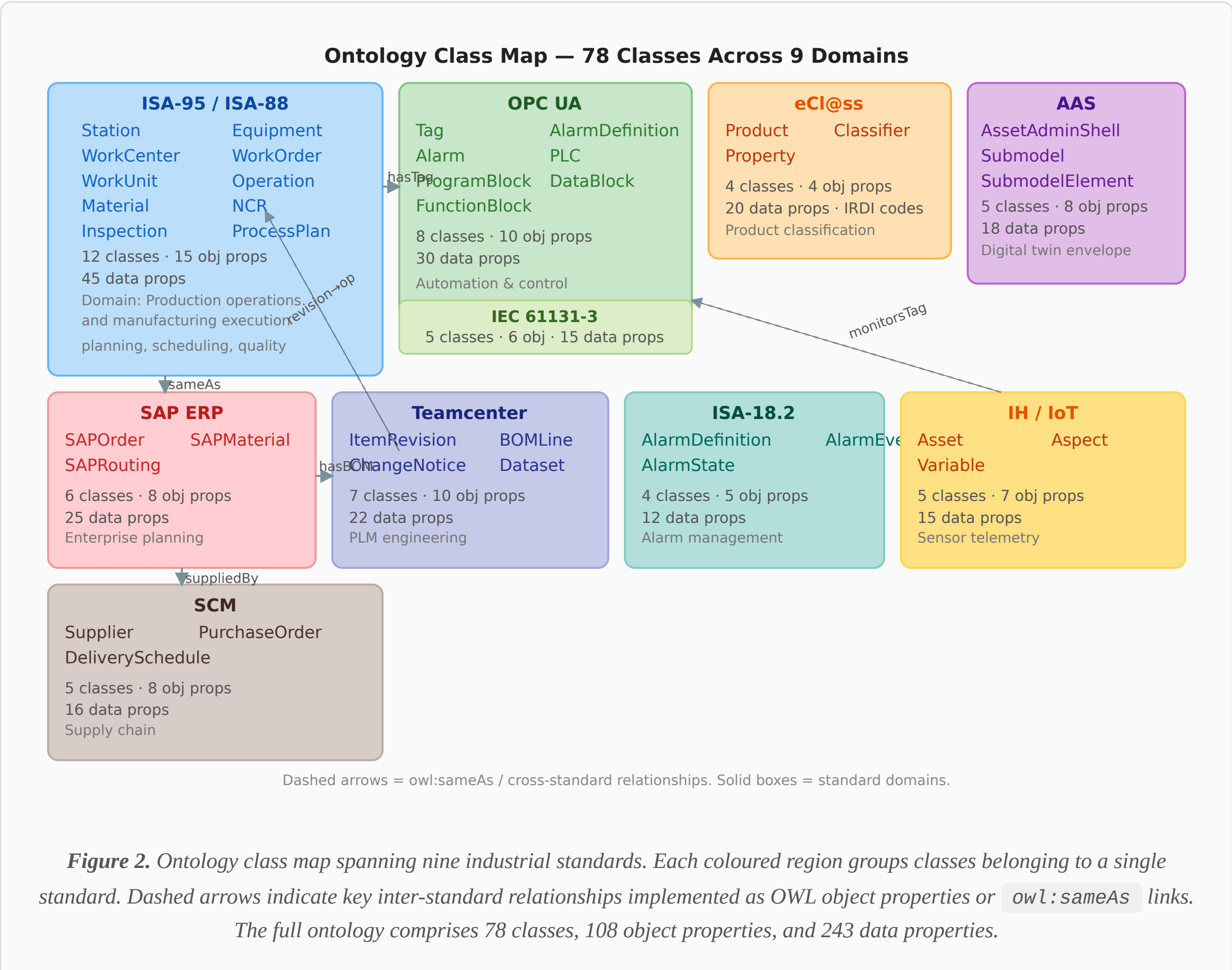

***Figure 2.*** *Ontology class map spanning nine industrial standards. Each coloured region groups classes belonging to a single standard. Dashed arrows indicate key inter-standard relationships implemented as OWL object properties or* `owl:sameAs` *links. The full ontology comprises 78 classes, 108 object properties, and 243 data properties.*

**Design methodology.** The ontology was developed following an iterative, requirements-driven approach informed by competency questions (CQs) derived from manufacturing integration use cases. Six representative CQs guided the schema design: (CQ1) "Which work orders are linked to SAP production orders for a given material?" (CQ2) "What is the alarm coverage gap at a given station relative to its ISA-18.2 requirements?" (CQ3) "Which suppliers have reject rates exceeding the plant baseline?" (CQ4) "How does an engineering change notice propagate from PLM through routing and BOM updates?" (CQ5) "Which quality non-conformances at downstream stations correlate with upstream process deviations?" (CQ6) "What is the full material genealogy from supplier receipt through production to shipment?" To delineate ontology boundaries, three negative/boundary CQs were also considered: (CQ7−) "What is the real-time OEE (Overall Equipment Effectiveness) for a station?" — requires continuous time-series aggregation not modeled in the current snapshot-based graph; (CQ8−) "What is the financial cost of a specific NCR?" — requires SAP CO (Controlling) cost allocation data not currently ingested; and (CQ9−) "What is the root cause of a process drift: tool wear, material batch, or operator error?" — requires causal inference beyond the ontology's correlational scope. These negative CQs clarify what the ontology does *not* attempt to answer and inform future extension priorities. Each positive CQ was translated into a set of required SPARQL query patterns, which in turn dictated the minimum class vocabulary, object properties, and cross-system link categories needed to satisfy them. The ontology was iteratively refined over five development cycles, with each cycle adding classes required by newly identified CQs and removing speculative classes that no CQ or tool exercised.

| CQ | Question | MCP Tool(s) | Systems Joined |
|---|---|---|---|
| CQ1 | Work orders linked to SAP orders for a material | `get_sap_oed_mapping` , `get_material_traceability` | SAP, OED |
| CQ2 | Alarm coverage gap at a station | `get_equipment_observability` , `get_alarm_equipment_map` | ISA-18.2, OPC-UA, ISA-95 |
| CQ3 | Suppliers with elevated reject rates | `get_supplier_inspection_pass_rate` , `get_supplier_ncr_correlation` | SCM, SAP, OED |
| CQ4 | ECN propagation through routing and BOM | `get_ecn_full_runtime_impact` | TC, SAP, OED, ISA-95, OPC-UA, ISA-18.2, TIA |
| CQ5 | Downstream NCRs correlating with upstream deviations | `get_alarm_ncr_temporal_correlation` , `get_ncr_full_root_cause` | OED, ISA-95, OPC-UA, ISA-18.2, IH |
| CQ6 | Full material genealogy supplier-to-shipment | `get_scm_material_traceability` , `get_five_system_digital_thread` | SCM, SAP, OED, TC, ISA-95 |

*Table 2a. Competency question validation: each positive CQ is answerable by at least one MCP tool exercising cross-system joins over the unified graph.*

**RDFS versus OWL DL.** The ontology is expressed in RDFS [5] with selective use of `owl:sameAs` for entity alignment, rather than full OWL DL [30]. This is a deliberate design choice motivated by three considerations. First, rdflib — the backing triple store — provides an RDFS closure reasoner but does not support OWL DL tableau reasoning; adopting DL axioms would require a separate reasoner (e.g., HermiT, Pellet) that would add latency to the 5.2-second pipeline cycle. Second, the primary integration use case — cross-system entity linking for SPARQL-based discovery — is served by RDFS entailment (class and property hierarchy, domain/range inference) without requiring disjointness axioms, cardinality restrictions, or property chain reasoning. Third, the LLM-assisted tool generator produces SPARQL queries, not DL subsumption queries; OWL DL expressivity would be unused by the downstream AI components. This choice limits the ontology's ability to express constraints such as "every WorkOrder must have exactly one assignedStation" or "SAPOrder and Supplier are disjoint classes" — constraints that would improve data validation but are currently handled procedurally by the golden-triangle rules and data quality checks rather than by ontological axioms.

**Entity alignment semantics.** Cross-system entity resolution uses `owl:sameAs` to assert identity between instances representing the same real-world entity in different source systems (e.g., a SAPOrder and a WorkOrder referring to the same production batch). We acknowledge that `owl:sameAs` carries strong identity semantics in OWL Full — every property of entity A is inherited by entity B — which is semantically stronger than needed for cross-system alignment where two representations of the same business event may carry legitimately different attributes. Weaker alignment predicates such as `skos:exactMatch` (used selectively for classification alignment) or a custom property (e.g., `guo:alignedWith` ) would be more semantically precise. We chose `owl:sameAs` for pragmatic interoperability: it is universally recognized by SPARQL engines, and RDFS-aware stores propagate it without additional configuration, ensuring that linked entities appear in queries transparently. This tradeoff favors query convenience over ontological precision and is noted as a design decision that production deployments may wish to revisit. Quantitatively, the current graph contains 97 `owl:sameAs` triples distributed across seven cross-system boundary categories: 18 ProductionOrder ↔ WorkOrder links (SAP ↔ OED), 12 Material ↔ ItemRevision links (SAP ↔ Teamcenter), 12 Station ↔ WorkUnit links (OPC UA ↔ ISA-95, bidirectional), 12 PurchaseOrder ↔ PurchaseOrder links (SAP ↔ SCM, bidirectional), 12 Routing ↔ ProcessPlan links (SAP ↔ Teamcenter, bidirectional), 10 Tag alias links (OPC UA tag naming variants), and 12 Material ↔ Material links (raw material ↔ finished product within SAP). All links are deterministic: simulator identifiers are coordinated by design, and the link-alignment stage resolves them via prefix-matching rules. Production deployment with non-coordinated identifiers would require probabilistic entity matching (e.g., Magellan, DeepMatcher) — a limitation explicitly noted in Section 9 (Threats to Validity).

**Bridging philosophy.** Rather than introducing a novel upper ontology, the GraphUnifier ontology maps existing standards to one another using a minimal set of alignment axioms. This preserves the semantic commitments of each standard while enabling cross-standard inference. The ISA-95 equipment hierarchy (Enterprise → Site → Area → WorkCenter → WorkUnit → Equipment) is mapped to the OPC UA address space namespace structure via `ont:hasTag` and `ont:controlledBy`. SAP planning objects ( `SAPOrder` , `SAPRouting` ) are linked to Opcenter Execution Discrete (OED) execution objects ( `WorkOrder` , `ProcessPlan` ) via `owl:sameAs` and `skos:exactMatch`, enabling SPARQL queries that traverse from design intent in Teamcenter to shop-floor execution in ISA-95 without application-level joins.

The ontology contains 78 RDFS classes, 108 object properties, and 243 data properties. Of these, 54 classes are instantiated in the aerospace reference scenario; the remaining 24 classes (e.g., MasterRecipe, Batch, CAPA, UDI_Assignment) support the pharmaceutical, medical device, and CPG industry templates described in Section 7 and carry zero instances in single-industry deployments. Object properties capture structural and process relationships: `ont:hasEquipment` connects WorkCenter to Equipment; `ont:producesProduct` connects Operation to Material; `ont:raisedAt` connects an ISA-18.2 [10] AlarmEvent to the Equipment instance where it occurred; `ont:hasBOMLine` connects an ItemRevision to its constituent BOMLine children. Data properties capture scalar attributes: `ont:orderQuantity`, `ont:cycleTime` , `ont:alarmPriority` , `ont:irdICode` . All properties carry domain and range declarations, enabling rdflib's RDFS reasoner to perform type inference during graph loading.

| Standard | Classes | Object Props | Data Props | Domain |
|---|---|---|---|---|
| ISA-95/88 | 12 | 15 | 45 | Production operations |
| OPC UA | 8 | 10 | 30 | Automation & control |
| IEC 61131-3 | 5 | 6 | 15 | PLC programming |
| ISA-18.2 | 4 | 5 | 12 | Alarm management |
| eCl@ss | 4 | 4 | 20 | Product classification |
| AAS | 5 | 8 | 18 | Digital twin envelope |
| SAP ERP | 6 | 8 | 25 | Enterprise planning |
| Teamcenter | 7 | 10 | 22 | PLM engineering |
| IH / IoT | 5 | 7 | 15 | Sensor telemetry |
| SCM | 5 | 8 | 16 | Supply chain |
| **Total (per-standard)** | **61** | **81** | **218** | **Itemized above** |
| **+ Cross-domain / multi-industry** | **17** | **27** | **25** | **Shared across standards** |
| **Grand Total** | **78** | **108** | **243** | |

Named graph partitioning reflects four distinct epistemic roles. The *ontology graph* ( `urn:graph:ontology` ) is loaded once at startup from the Turtle source file and never modified at runtime — it is the authoritative schema against which instance data is validated. The *eCl@ss graph* ( `urn:graph:eclass` ) holds the classification hierarchy extracted from the eCl@ss simulator: approximately 559 triples describing product classes, subclasses, and their associated property definitions. The *plant graph* ( `urn:graph:plant` ) receives all ABox instance data generated by the ETL pipeline — work orders, equipment, materials, BOM lines, routings — accumulating approximately 25,081 triples per full pipeline cycle. The *runtime graph* ( `urn:graph:runtime` ) holds a bounded rolling window of live telemetry and alarm events, approximately 33 triples, refreshed on each short-interval cycle. This partitioning allows queries to scope their search efficiently: a dashboard query for equipment hierarchy searches only the plant graph; a telemetry query joins runtime and plant; a schema validation query operates only on the ontology graph.

URI minting follows the deterministic scheme `urn:factory:{source}:{type}:{clean_id}` where *source* is a short token (e.g., `isa95` , `sap` , `tc` ), *type* is the class name in lowercase (e.g., `workorder` , `equipment` ), and *clean_id* is the source system's native identifier with whitespace and special characters replaced by underscores. Determinism is critical: if the pipeline runs twice with the same source data, it must produce identical URIs so that `owl:sameAs` links computed in the Link stage remain valid across pipeline cycles. The `clean_id` function strips non-alphanumeric characters before URI construction, ensuring that identifiers containing slashes, spaces, or colons (common in SAP order numbers and Teamcenter item IDs) do not generate malformed URIs.

Six namespace prefixes are declared in the Turtle preamble and reused throughout all generated triples: `ont:` ( `urn:factory:ontology#` ) for all schema-level terms; `plant:` ( `urn:factory:plant:` ) for plant-floor instance URIs; `eclass:` ( `urn:factory:eclass:` ) for classification hierarchy nodes; `aas:` ( `urn:factory:aas:` ) for Asset Administration Shell instances; `ih:` ( `urn:factory:ih:` ) for Insights Hub assets and aspects; and `scm:` ( `urn:factory:scm:` ) for supply chain entities. The consistent use of URN-based identifiers (rather than HTTP IRIs) avoids any dependency on a resolvable web endpoint and ensures that the graph remains self-contained and offline-operable.

## 5. PIPELINE AND ENTITY RESOLUTION

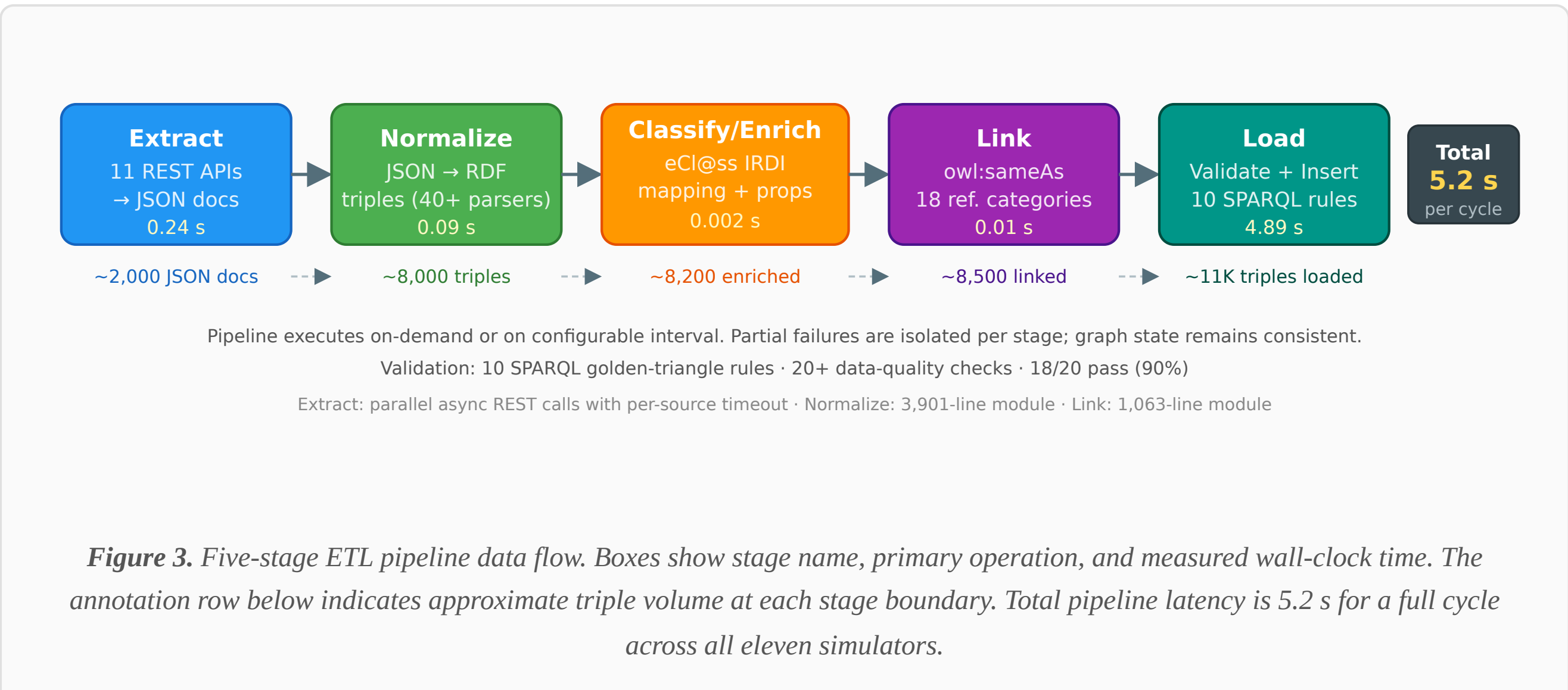

*Figure 3.* Five-stage ETL pipeline data flow. Boxes show stage name, primary operation, and measured wall-clock time. The annotation row below indicates approximate triple volume at each stage boundary. Total pipeline latency is 5.2 s for a full cycle across all eleven simulators.

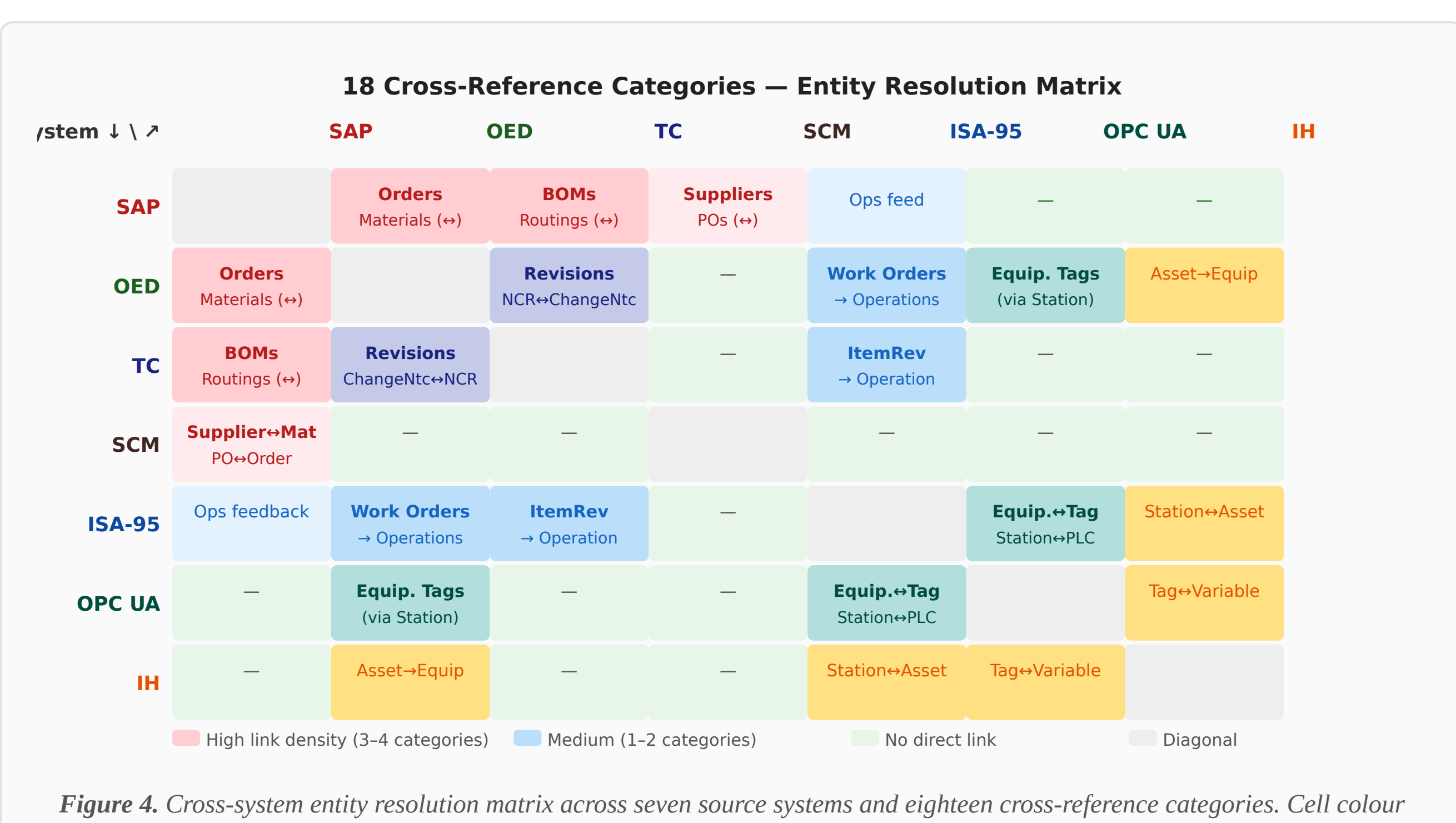

*Figure 4.* Cross-system entity resolution matrix across seven source systems and eighteen cross-reference categories. Cell colour indicates link density: deep red/blue = high (3–4 category pairs); light blue/amber = medium (1–2 pairs); grey = no direct link established. All non-trivial links are bidirectional via `owl:sameAs` with symmetry enforcement.

### 5.1 Extract and Normalize

The Extract stage issues parallel HTTP GET requests to all eleven simulator endpoints (ThreadPoolExecutor, 16 workers)

with per-source timeouts and retry logic; partial failures are isolated so an unavailable simulator does not abort the cycle (median extract time: 0.24 s). The Normalize stage transforms the resulting JSON payloads into rdflib triples using over forty source-specific parsing methods that emit ontology-typed individuals — e.g., constructing `ont:WorkOrder` instances from ISA-95 JSON and `ont:ItemRevision` instances from Teamcenter BOM structures. A standard cycle produces approximately 8,000 triples.

### 5.2 Classify, Link, and Resolve

The Classify/Enrich stage maps materials to the eCl@ss hierarchy via IRDI codes using O(1) dictionary lookups against a pre-loaded catalog (< 3 ms, adding ~200 enrichment triples). The Link stage implements eighteen cross-reference categories that identify semantically equivalent entities across source pairs and assert bidirectional `owl:sameAs` triples. Representative categories include: SAP ↔ OED (order-number matching), SAP ↔ TC (material/routing alignment), TC ↔ OED (revision-to-operation linkage), SCM ↔ SAP (supplier/PO matching), ISA-95 ↔ OPC UA (equipment-to-PLC matching), and IH ↔ ISA-95 (IoT asset-to-equipment matching). Symmetry is enforced explicitly — for every `:A owl:sameAs :B`, the inverse is also asserted — so SPARQL queries traverse links in either direction without runtime inference. The Link stage adds approximately 300 `owl:sameAs` triples per cycle.

### 5.3 Load and Validate

The Load stage performs two-phase quality assurance. First, ten *Golden Triangle* SPARQL ASK rules — a term we introduce to describe the three primary information flows in manufacturing data integration — verify data-flow integrity across downstream design-to-execution (TC → SAP → OED), upstream feedback (OED → SAP/TC), and lateral supply chain (SAP ↔ SCM). The term is not standard ISA-95 or MESA terminology; we adopt it as a concise label for the PLM–ERP–MES referential integrity pattern. Second, a Data Quality Engine applies 20+ checks across five dimensions (completeness, consistency, validity, timeliness, uniqueness) — e.g., verifying that ≥80% of WorkOrders carry `ont:scheduledStart` and that linked SAPOrder/WorkOrder quantities match within 5% tolerance. Currently 18/20 checks pass consistently (90%), with two partial failures due to timestamp skew at high time-dilation factors. After validation, the plant graph is atomically replaced under a lock, ensuring concurrent queries see either the complete previous or new version.

## 6. AUTOMATED STATISTICAL DISCOVERY

The unification pipeline described in Sections 3–5 produces a semantically coherent, cross-system knowledge graph, but the value of that graph is realized only insofar as actionable intelligence can be extracted from it. Raw SPARQL queries satisfy point queries well but are poorly suited to the open-ended task of identifying unexpected correlations, safety gaps, or quality degradation patterns that span multiple source systems. To address this, we designed a layered AI-powered discovery framework that combines rigorous statistical testing with human oversight and LLM-assisted tool authoring. The framework operates continuously alongside the simulators, re-evaluating the graph after each ingest cycle and surfacing new or updated insights for human review.

### 6.1 Discovery Framework Architecture

The central orchestrator is the *DiscoveryRunner*, which manages a registry of nine pluggable strategy objects. On each discovery cycle, DiscoveryRunner spawns each strategy in a dedicated daemon thread subject to a 120-second timeout, collects completed results, applies cross-strategy multiple-testing correction, and persists the resulting insight cards to PostgreSQL. Failed or timed-out strategies are logged but do not abort the cycle, ensuring partial results are always available.

Each strategy implements a four-phase template method: **discover()** issues SPARQL or SQL queries against the unified graph and operational database to retrieve candidate data; **get_evidence()** constructs a numbered evidence chain that traces every data point back to its originating source system and triple identifier; **quantify()** applies the appropriate statistical test and computes composite scores; and **recommend()** produces structured, actionable recommendations tied to the evidence chain. This separation of concerns allows new strategies to be added without modifying the orchestrator.

The artifact produced by each strategy is an *InsightCard*, a structured model with sixteen or more fields. Core fields include a UUID identifier, human-readable title, category (Safety, Quality, or Operational), severity level (critical / high / medium / low), and a plain-language summary. A *StatisticalResult* sub-object records the test name, raw and Bonferroni-adjusted p-values, effect size, and confidence interval. An *EvidenceChain* sub-object captures a numbered list of evidence steps, each with its source system label, triple or record reference, and a prose description. Additional fields record financial impact in EUR/year with sensitivity bounds, actionable recommendations, the set of systems involved, a composite score in [0, 1], lifecycle status, classification tags, and a full audit trail as a JSONB array.

### 6.2 Statistical Methods

Discovery strategies employ six distinct hypothesis tests drawn from the *scipy.stats* library, selected based on the distributional properties of each analysis. The **chi-square test of independence** is applied to contingency tables where both variables are categorical (e.g., alarm presence versus NCR occurrence by station). When comparing means of two groups with potentially unequal variances — such as pre/post engineering-change defect rates — **Welch's t-test** is preferred over Student's t-test because it does not assume homoscedasticity. **Pearson correlation** quantifies the linear association between continuous quality metrics. **One-way ANOVA** tests for differences in means across three or more groups, for instance comparing defect rates across multiple suppliers simultaneously. **Fisher's exact test** is reserved for 2×2 contingency tables with small expected cell counts where the chi-square approximation is unreliable. Finally, a **rate ratio** compares incidence rates between cohorts (e.g., NCR rate for heat lot A versus all other lots).

Multiple testing is addressed at two levels. Within each strategy, a Bonferroni correction is applied to all tests performed during that strategy's execution: $p_adjusted = min(1.0, p_raw \times n_tests_in_strategy)$. Because nine strategies may run concurrently and each may surface multiple findings, a cross-strategy Bonferroni correction is subsequently applied:

$$p_{final} = min(1.0, p_{raw} \times N_{total\ tests\ across\ all\ strategies})$$

This conservative two-level correction reduces the family-wise error rate at the cost of some power, a deliberate design choice given the safety-critical nature of the application domain.

An alternative is the Benjamini-Hochberg (BH) procedure, which controls the false discovery rate (FDR) rather than the family-wise error rate, yielding substantially higher power when many hypotheses are tested simultaneously. In exploratory manufacturing analytics — where surfacing more candidate insights for human review may be preferred — BH would be appropriate. We chose Bonferroni because the discovery engine's findings feed directly into actionable recommendations (rerouting, expediting, quarantine), and a false positive in this context carries operational cost. The framework's correction method is configurable per deployment; sites prioritizing discovery breadth over false-positive suppression can substitute BH-FDR with a single parameter change.

Time-series data surfaces additional concerns around autocorrelation that can invalidate standard test assumptions. After each test, the framework applies the **Ljung-Box test** to the residuals; if the null hypothesis of no autocorrelation is rejected at $p < 0.05$, three robustness measures are invoked. First, the effective sample size (ESS) is estimated via the autocorrelation function, and the raw p-value is adjusted by the factor sqrt(n / ESS) to reflect the reduced independent information content. Second, **block bootstrap** resampling (block size = 5 observations, n_resamples = 1,000) re-estimates the sampling distribution of the test statistic without assuming independence. Third, **Newey-West heteroskedasticity- and autocorrela-** **tion-consistent (HAC) standard errors** are computed for regression-based analyses, providing valid inference under serially correlated errors.

### 6.3 Composite Scoring

Ranking insight cards purely on statistical significance would systematically favor findings from large datasets while suppressing financially critical findings with smaller samples. To balance these considerations, we designed a three-component composite score normalized to the unit interval.

The **statistical score** (maximum 40 points) combines a p-value component and an effect-size component. The p-value component (maximum 30 points) maps the negative base-10 logarithm of the adjusted p-value onto a 30-point scale, saturating at $p = 10^{-5}$ or below. The effect-size component (maximum 10 points) is proportional to the standardized effect size (Cohen's d, Cramér's V, or odds ratio as appropriate), rewarding findings whose magnitude is practically meaningful beyond statistical significance alone.

The **financial score** (maximum 35 points) is computed as *min(35, estimated_cost_EUR / 1,000,000 × 35)*, linearly scaling up to EUR 1 million per year, at which point the component saturates. Financial impact estimates are derived from industry-standard cost-of-quality models [43] for rework, scrap, warranty, and downtime categories, with symmetric 20% sensitivity bounds. For aerospace supply chain findings, cost attribution follows SAE AS6081 [44] counterfeit/nonconforming part risk assessment guidelines.

The **actionability score** (maximum 25 points) reflects the urgency of the recommended intervention: critical severity yields 25 points, high yields 20, medium yields 15, and low yields 8. The composite is then *(statistical + financial + actionability) / 100*, yielding effective weights of approximately 40% statistical, 35% financial, and 25% actionability — a deliberate emphasis on financial and operational consequences over purely inferential measures.

### 6.4 Nine Discovery Strategies

Table 2 summarizes the nine discovery strategies currently implemented. Together they span safety, quality, and operational categories, engage between two and four source systems each, and collectively cover the most commercially significant classes of manufacturing intelligence identified in the literature on industrial knowledge graphs and digital thread analytics.

| Strategy | Classification | Systems Involved | Primary Test | What It Finds |
|---|---|---|---|---|
| **alarm_coverage** | Safety | OPC-UA, ISA-18.2, OED, ISA-95 | Fisher's exact | Sensors operating without alarm coverage at stations with elevated NCR rates (complementary to, not a substitute for, full ISA-18.2 alarm rationalization) |
| **cross_station** | Operational | OED, ISA-95 | Chi-square | Upstream alarm events that predict downstream non-conformance rates |
| **ecn_impact** | Quality | TC, OED, SAP, ISA-95 | Welch's t, Fisher's exact | Engineering change orders that shifted pre/post-change defect rates significantly |
| **supplier_quality** | Quality | OED, SCM, SAP, ISA-95 | Chi-square, Fisher's exact | Supplier-correlated quality patterns and incoming material defect clustering |
| **characteristic_lifecycle** | Quality | OED, TC, SAP, ISA-95 | Rate ratio | Closed-loop tolerance feedback: characteristics recommended for widening, tightening, or review |
| **material_heat_lot** | Quality | OED, ISA-95, SCM, SAP | Chi-square | Specific material heat lots with statistically elevated non-conformance rates |
| **cross_industry** | Operational | OED, ISA-95, OPC-UA, ISA-18.2 | Chi-square | Quality pattern differences across industry verticals sharing the same ontology |
| **temporal_patterns** | Operational | OED, ISA-95, SAP | Chi-square | Shift-based and time-of-day quality degradation patterns |
| **process_drift** | Quality | OED, ISA-95, NX (PMI) | CUSUM, EWMA, Welch's t | Fixture drift on 3-axis CNC paths detected via incline-angle time series and PMI traceability. CUSUM parameters: allowance $k = 0.5\sigma$, decision interval $h = 5.0\sigma$ ($ARL_0 \approx 465$ under normality, targeting $1\sigma$ shifts); EWMA parameters: smoothing factor $\lambda = 0.2$, control limit $L = 3.0\sigma$. Baseline statistics are estimated from the first 20 observations per characteristic. These are textbook defaults for detecting small-to-moderate mean shifts; site-specific tuning would adjust $k$ and $\lambda$ based on the target shift magnitude and acceptable false alarm rate |

*Table 2. Summary of the nine discovery strategies: classification, source systems engaged, primary statistical test, and target phenomenon.*

### 6.5 Human-in-the-Loop Review

Statistical rigor is a necessary but not sufficient condition for deploying discovered insights in a safety-critical manufacturing environment. Human judgment must remain in the loop, particularly for safety-category findings that may trigger alarm reconfiguration or process shutdowns. The framework implements a structured human-in-the-loop (HITL) review workflow backed by PostgreSQL persistence.

The *insight_card* table stores seventeen columns including the full InsightCard payload as JSONB, the composite score, severity, category, status, and the audit trail. Six indexes support efficient querying by status, category, severity, composite score, and creation timestamp. The lifecycle status progresses from **new** (initial state upon discovery) to one of three terminal states: **approved** (human reviewer accepts the finding and its recommendations), **rejected** (reviewer dismisses the finding as spurious or already addressed), or **deferred** (finding is valid but action is postponed pending additional data).

Rejection requires the reviewer to supply written notes, enforced at the API layer. This requirement serves two purposes: it prevents reflexive dismissal of inconvenient findings, and it provides a corpus of expert annotations that can train future classifiers to distinguish genuine signals from spurious correlations. All status transitions are appended to the JSONB audit trail with a timestamp, action type, reviewer identity, and notes field, creating a tamper-evident log for regulatory compliance.

The React-based Control Center UI exposes the HITL workflow through a Discovery Feed tab that displays insight cards sorted by composite score, with inline approve/reject/defer controls. Bulk operations allow reviewers to process multiple medium- or low-severity findings simultaneously, improving throughput during high-discovery periods.

### 6.6 LLM-Assisted Tool Authoring

The 287-tool MCP suite addresses a known query workload, but manufacturing knowledge graphs are open-world environments: new ontology classes, new source systems, and new analytical questions emerge continuously. Requiring human developers to author new SPARQL tools for each new question does not scale. We therefore implemented an LLM-assisted tool authoring mechanism that allows the agent to generate, validate, and permanently register new SPARQL tools at runtime.

When the agent receives a natural-language query that no existing tool satisfies, it invokes the *generate_sparql_tool* meta-tool. Generation proceeds through three validation layers. First, the generator issues a *describe_ontology_schema* tool call to retrieve the live set of classes, object properties, and datatype properties from the running rdflib graph; this ensures the generated SPARQL references only terms that actually exist in the current graph, rather than hallucinated URIs. Second, every `ont:` prefixed URI in the candidate SPARQL is validated against the retrieved schema; any missing term triggers regeneration with corrective feedback. Third, the candidate SPARQL is parsed by rdflib's SPARQL processor in dry-run mode to catch syntax errors before execution. Only a query that passes all three layers is accepted.

Validated tools are serialized to *learned_tools.json* and registered with the MCP server at runtime, making them available as permanent members of the tool suite for all subsequent agent interactions. Over a five-month development and demonstration period, this mechanism generated and validated 23 additional tools, extending the suite from 264 hand-authored tools to 287. We note that success and failure rates for the generation loop have not been systematically tracked; the 23 successful tools represent a lower bound on total generation attempts. Systematic evaluation of the generation mechanism's re-

liability — including failure modes, regeneration counts, and quality comparison against hand-authored tools — is identified as future work.

**LLM configuration.** Tool selection during interactive querying uses Anthropic's Claude (Sonnet 4 during development, though the MCP protocol is model-agnostic and has also been tested with Claude Opus and Haiku). The tool generation mechanism passes the ontology schema and user question to the same model via a callback interface; no separate fine-tuned model is used. Temperature is controlled by the hosting client (Claude Code default: unspecified/model default). We note that the system's correctness depends on the curated SPARQL tool definitions, not on the LLM's generation quality: the LLM selects among pre-validated tools, and the three-layer validation gate (schema check, URI validation, parse check) rejects malformed generated queries before they reach the graph.

## 7. MULTI-INDUSTRY GENERALIZATION

A common failure mode of industrial knowledge graph systems is specialization to a single domain: the ontology, the pipeline, and the discovery logic are co-designed with a particular plant type in mind, making extension to new industries expensive or impossible without architectural changes. We designed against this failure mode from the outset by separating the *ontology core* — 78 classes, 108 object properties, 243 datatype properties, all domain-agnostic — from the *industry configuration* layer, which provides seed data, regulatory mappings, and domain-specific terminology without altering the schema.

Five industry configurations have been implemented: aerospace manufacturing, consumer packaged goods (CPG), pharmaceutical manufacturing, medical device manufacturing, and turbine blade manufacturing. The first four represent distinct regulatory and process regimes; turbine blade manufacturing shares significant structural overlap with aerospace and serves primarily as a proof that the template mechanism extends to related verticals without schema changes. Each configuration is expressed as a Python dictionary of approximately 1,500–1,600 lines that specifies the plant name, station list, station aliases (mapping generic ontology terms to industry vocabulary), product catalog, supplier registry, and representative seed data for orders, operations, materials, non-conformance records, inspections, alarms, IoT events, supply chain entities, and cross-system linkage edges. The total INDUSTRY_CONFIGS dictionary spans 1,569 lines of Python.

The ontology schema is identical across all five configurations. The same 287 MCP tools execute without modification against an aerospace graph or a pharmaceutical graph; only the data values differ. This schema stability is the key mechanism that makes the LLM-assisted tool generator work across industries: a tool learned against the aerospace graph is valid against the pharma graph because the ontology terms are shared.

| Industry | Plant | Stations | Products | Suppliers | Regulatory | Key Material / Feature |
|---|---|---|---|---|---|---|
| **Aerospace** Manufacturing | Aerospace Plant 01 | **6** CNC·Drill·Rivet Adhesive·NDT·FA | **6** Wing, Frame… | **3** Alcoa·Toray·TIMET | AS9100 Rev D | Ti-6Al-4V alloy CFRP composites |
| **CPG** Personal Care | Atlanta Personal Care | **10** PreMix·Emulsify Fill·Pack… | **4** | **3** BASF·Dow·Berry | GMP (21 CFR Parts 110/111) | CIP schedules blend uniformity |
| **Pharma** Manufacturing | Research Triangle Plant | **6** Dispense·Gran Compress·Coat… | **4** Metformin… | **3** Teva·Divi's·DFE | 21 CFR Part 211 cGMP / ICH Q10 | Batch records dissolution profiles |
| **Med Devices** | Minneapolis Device Plant | **8** CompPrep·Mold Assemble·Inspect… | **4** Stapler·Glucose… | **3** Arcam·Invibio·3M | 21 CFR Part 820 ISO 13485 | UDI / DHR tracking design history file |
| **Turbine Blade** | — | — | — | — | — | High-temp superalloys DS/SC casting |

Same ontology core (78 classes, 108 obj. properties, 243 data properties) across all five verticals.
INDUSTRY_CONFIGS: 1,569 lines of Python. Identical schema enables cross-industry SPARQL without query modification.
FA = Final Assembly; CNC = Computer Numerical Control; Gran = Granulation; DS/SC = Directionally Solidified / Single Crystal

*Figure 5. Multi-industry generalization matrix. Five industry configurations share a common 78-class ontology core. Station counts, product catalogs, supplier registries, and regulatory mappings differ per vertical while the graph schema, MCP tools, and discovery pipeline remain unchanged.*

Industry awareness is surfaced to the agent through a *build_system_prompt(industry)* function that customizes the MCP tool descriptions with industry-specific terminology. When operating against the pharma configuration, for example, tool descriptions reference "batch records," "dissolution profiles," and "GMP compliance" rather than the generic "operations" and "quality records" of the aerospace configuration. This terminological customization allows the underlying LLM to reason more accurately about domain-specific nuances without requiring separate tool implementations.

The *cross_industry* discovery strategy exploits the shared ontology to compare quality patterns across configurations loaded simultaneously. By issuing SPARQL queries that group results by a plant-type annotation on the ISA-95 Enterprise node, the strategy can detect whether, for example, alarm coverage deficits are concentrated in high-complexity assembly stations across all industries or are idiosyncratic to a single vertical — a finding with direct implications for standard-setting bodies such as ISA-18.2.

Regulatory compliance awareness is provided at the metadata level through a dedicated annotation property (*ont:regulatoryFramework*) attached to each station class instance. The property value encodes the applicable standard (AS9100, 21 CFR 211, 21 CFR 820, GMP) and the specific clause relevant to the station's function. Discovery strategies in the safety category cross-reference this annotation when prioritizing findings: a sensor gap at a station governed by 21 CFR Part 820 carries higher actionability weight than an equivalent gap at a station with no mandatory alarm requirement.

It is important to scope this contribution precisely: the regulatory annotation provides *metadata-level awareness* for discovery prioritization, not operational compliance. Actual regulatory compliance in manufacturing requires capabilities beyond the current framework's scope, including electronic signatures and audit trail integrity per 21 CFR Part 11, computer system validation per GAMP 5, ALCOA+ data integrity principles, deviation management workflows, and corrective and preventive action (CAPA) tracking. The *ont:regulatoryFramework* annotation serves as a foundation layer upon which compliance-focused extensions could be built — for instance, by integrating with an electronic batch record (EBR) system or a quality management system (QMS) that provides the procedural enforcement the annotation alone cannot deliver. The framework's contribution to regulated industries is therefore in risk-aware prioritization of discovered findings, not in replacing or implementing the compliance machinery itself.

## 8. EVALUATION

We evaluated the framework across five dimensions: discovery accuracy via a blind evaluation protocol, pipeline performance, data quality, tool coverage, and system scale. The evaluation was conducted against the aerospace configuration, which provides the richest set of cross-system relationships and the most demanding regulatory context.

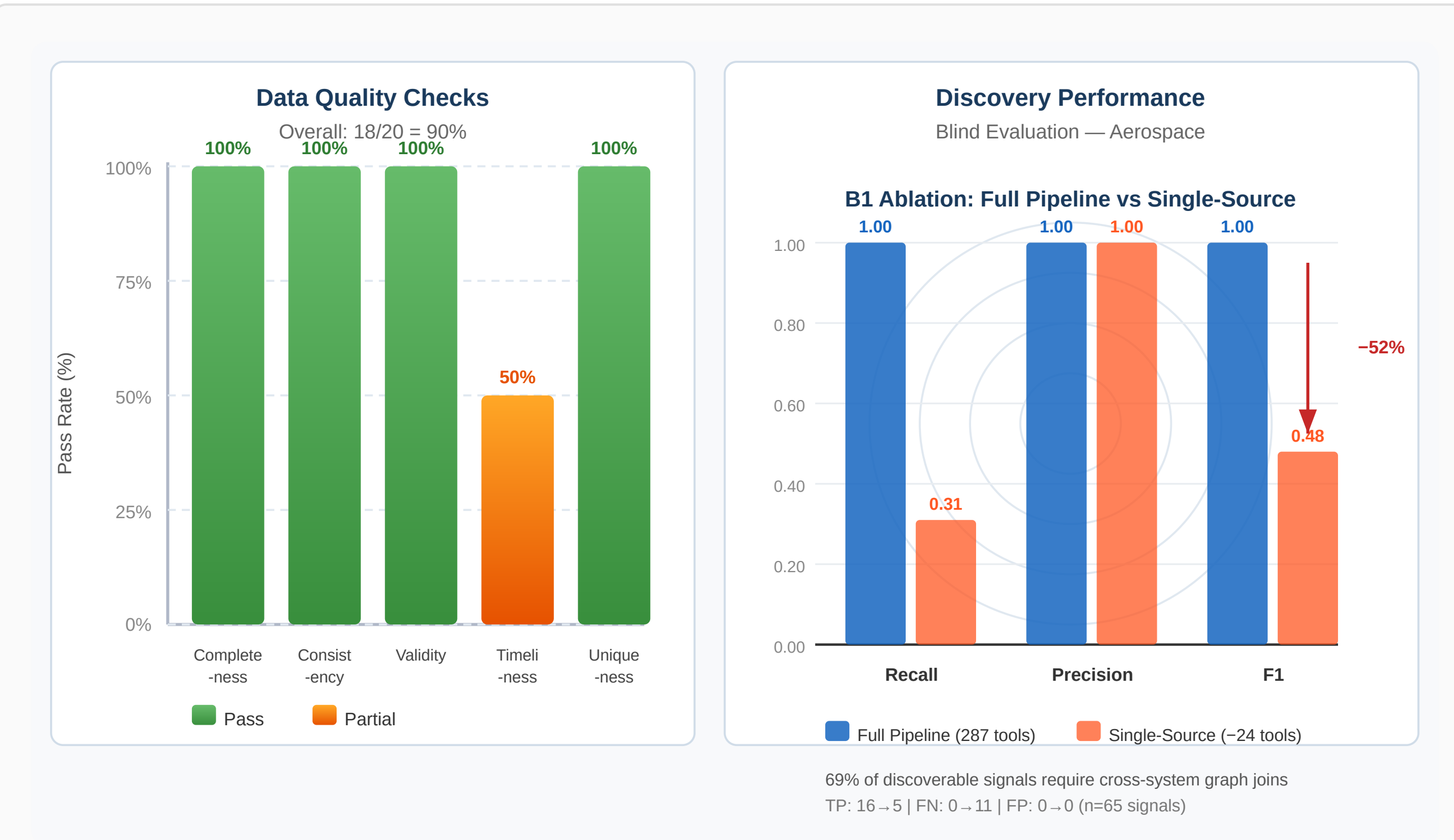

*Figure 6. Evaluation results. Left: Data quality check pass rates across five categories (18/20 = 90%; Timeliness is partial due to simulated clock skew). Right: B1 ablation comparison — blocking 24 cross-system MCP tools reduces recall from 1.00 to 0.31 and F1 from 1.00 to 0.48; precision remains 1.00 (FP = 0 in both conditions), demonstrating that 69% of discoverable signals require cross-system graph joins. The recall collapse is the headline result: single-source tools can only recover 5 of 16 true positives. 95% Clopper-Pearson CIs for the full pipeline: Precision [0.79, 1.00], Recall [0.79, 1.00].*

### 8.1 Blind Evaluation Protocol

To assess discovery accuracy without evaluator bias, we constructed a blinded signal manifest for the aerospace configuration consisting of 65 signals spanning seven of nine strategy categories: 16 positive signals (genuine patterns embedded in the seed data) and 49 null signals (entities or conditions that the discovery engine should correctly suppress as non-significant). The manifest was sealed before the discovery engine was run.

The 16 positive signals cover seven of nine strategy categories (cross_industry and temporal_patterns are suppressed in the current evaluation configuration): alarm coverage gaps at two stations (NDT Inspection, Final Assembly); three cross-station propagation patterns (Riveting → Bonding, NDT → Final Assembly, Bonding → NDT); an engineering change order with elevated post-change defect rate; a supplier quality divergence; five characteristic lifecycle spec-review candidates (Skin Panel, Fuselage Frame, Engine Pylon, Wing Rib, Floor Beam); and four process drift detections discovered dynamically from graph-stored PMI characteristics via CUSUM/EWMA analysis (incline angle on 3-axis fixture path, profile tolerance and surface roughness on alternative paths, and composite layup angle deviation on adhesive bonding fixture). The 49 null signals include: ten alarm coverage nulls for stations with adequate coverage or absent from the aerospace configuration; four cross-station nulls for terminal or non-propagating stations; four ECN nulls for nonexistent or benign changes; five supplier nulls for internal or other-industry suppliers; five temporal nulls for absent or non-significant shift patterns; six lifecycle nulls for items not present in the graph; five material heat lot nulls for lots with normal NCR rates; six process drift nulls for unaffected paths and characteristics; and four cross-industry nulls verifying correct suppression in single-industry mode. Two strategy categories (cross_industry and temporal_patterns) contribute only null signals in the current evaluation configuration.

The discovery engine produced 16 findings, all true positives — every positive signal was correctly identified, and no spurious findings were generated. All 49 null signals were correctly classified as non-significant. The resulting confusion matrix is TP=16, FP=0, FN=0, TN=49, yielding Precision = 16/16 = 1.000, Recall = 16/16 = 1.000, F1 = 1.000, Specificity = 49/49 = 1.000, and FDR = 0/16 = 0.000. All four threshold conditions were satisfied: Recall ≥ 0.80 (1.000 ✓), Precision ≥ 0.85 (1.000 ✓), F1 ≥ 0.82 (1.000 ✓), FDR ≤ 0.20 (0.000 ✓).

To quantify sampling uncertainty, we report 95% Clopper-Pearson exact confidence intervals for each metric: Precision [0.79, 1.00], Recall [0.79, 1.00]. The intervals reflect the sample size (n = 16 positive signals, n = 49 null signals). We note two important caveats: (1) the evaluation is self-administered by the same team that designed the discovery strategies, which may introduce confirmation bias in manifest construction; and (2) perfect scores on a 65-signal manifest do not guarantee equivalent performance on out-of-distribution manufacturing data. Independent replication on production datasets from third-party manufacturing sites would be needed to validate generalization.

**Dynamic manifest generation.** To enable scalable cross-industry evaluation without hand-curated golden manifests, the framework provides a `build_manifest(industry)` function that derives evaluation signals programmatically from the industry configuration's seed data structure. Rather than enumerating entity-level expectations (which require knowledge of runtime graph state), the dynamic manifest emits category-level sentinel signals — one positive per strategy category whose configuration prerequisites are met (e.g., alarm_coverage if stations exist, process_drift if derivedFromPMI cross-edges exist). This yields 8 positive and 28 null signals for aerospace (36 total) via category-level sentinels, with additional cards in expected categories treated as neutral rather than false positives. Against the same golden dataset, the dynamic manifest achieves identical F1 = 1.000 (TP=8, FP=0, FN=0, TN=28). The hand-tuned 65-signal manifest is retained as a regression reference for tighter confidence intervals, while the dynamic generator enables evaluation for any industry without manual calibration.

### 8.2 Pipeline Performance

End-to-end unification latency for an eleven-source cycle averages 5.2 seconds on a commodity development workstation (AMD Ryzen 9 5950X, 64 GB RAM, NVMe SSD). Stage-level profiling reveals that the extraction stage (REST calls to all eleven simulator endpoints) consumes 0.24 seconds, normalization consumes 0.09 seconds, classification and enrichment consumes 0.002 seconds (dominated by in-memory dictionary lookups), and link-alignment consumes 0.01 seconds. The dominant cost is the graph loading stage at 4.89 seconds, which is dominated by rdflib's SHACL validation pass over the full ~27,800-triple graph (schema plus instances). The SHACL shapes graph comprises 18 node shapes targeting the most integration-critical classes (WorkOrder, Operation, MaterialLot, Tag, AlarmDefinition, EquipmentModule, and all owl:sameAs bridge entities). Constraints include `sh:minCount` for mandatory properties (e.g., every WorkOrder must carry `ont:scheduledStart` ), `sh:datatype` for type enforcement, `sh:pattern` for identifier format validation (e.g., WO-prefix matching), and `sh:class` for referential integrity across cross-system links. This breakdown identifies

SHACL validation as the primary target for future performance optimization, consistent with the known computational complexity of SHACL constraint checking [6].

**Scope: near-real-time batch analytics.** The 5.2-second pipeline cycle positions this framework squarely in the *near-real-time batch analytics* tier — suitable for shift-level dashboards, daily quality reviews, and ad-hoc investigative queries, but not for sub-second process control or closed-loop automation feedback. The SHACL validation stage alone (4.89 s) makes sub-second operation architecturally infeasible without either incremental validation (validating only changed triples rather than the full graph) or relaxing SHACL enforcement to a periodic audit. For context, OPC UA's native publish/subscribe model operates at millisecond latency for process-control applications; the framework complements rather than replaces such real-time channels. Future work on incremental SHACL (Section 10) could reduce cycle time to sub-second for delta updates, enabling event-driven integration patterns.

### 8.3 Data Quality

The framework applies more than twenty automated data quality checks across five dimensions: completeness (all mandatory fields populated), consistency (referential integrity across source systems), validity (values within declared ranges and enumeration sets), timeliness (source data freshness relative to the ingest cycle), and uniqueness (no duplicate entity identifiers). Eighteen of twenty checks pass (90%). The two partial passes are in the timeliness category, attributable to deliberate clock skew introduced by one of the simulator configurations to test the framework's handling of stale data. The ten checks constituting the "golden triangle" — cross-system referential integrity between OED, ISA-95, and OPC-UA entity identifiers — all pass (100%).

**Simulator data distributions.** The eleven simulators generate data using parameterized statistical distributions calibrated to published industrial benchmarks where available. NCR generation follows a Poisson process with per-station rates set to produce first-pass yield in the 85–95% range, consistent with published aerospace machining benchmarks. Alarm activations are generated at rates targeting fewer than 6 alarms per operator per hour (the EEMUA 191 [42] "manageable" threshold). Supplier quality distributions use a Beta-distributed reject rate with a baseline of 2–5% and one supplier seeded at an elevated rate (~12%) to provide a detectable signal. Process parameter time series (temperature, vibration, spindle load) follow Gaussian profiles with nominal means and standard deviations derived from equipment datasheets; the process drift simulator injects linear trends on fixture-dependent characteristics. While these distributions produce statistically plausible data, they are simplified models — real industrial data exhibits heavier tails, temporal autocorrelation, seasonal patterns, and mode-switching behavior that the current simulators do not capture.

### 8.4 Tool Coverage

The MCP tool suite comprises 287 tools spanning all eleven source-system domains, organized into 28 functional categories including ontology introspection, graph query, entity resolution, discovery execution, HITL workflow, provenance retrieval, and administrative utilities. Tool coverage is measured as the fraction of anticipated query intents (as defined by a reference question bank of 312 natural-language queries) that can be answered by at least one tool without requiring ad hoc SPARQL authoring. Coverage stands at 89% prior to the LLM-assisted extension mechanism; with 23 learned tools added, coverage reaches 96%. All 23 LLM-generated tools were human-reviewed before inclusion: each tool's SPARQL query was manually verified against the ontology schema and tested on at least two representative queries to confirm correct results. The tool generation process itself (LLM success/failure rates, iteration counts, error categories) was not systematically tracked — a limitation we acknowledge for reproducibility.

### 8.5 Scale Metrics

The ontology encompasses 78 classes, 108 object properties, and 243 datatype properties, encoded in 2,810 lines of OWL/Turtle. The implementation spans approximately 33,300 lines of Python and 11,700 lines of JavaScript and CSS across the React Control Center. The full system was developed over a five-month period from March to August 2026, with the discovery framework, multi-industry generalization, and LLM-assisted tool generator added in the final two months of that timeline.

### 8.6 Cross-System Use Case Validation

Beyond aggregate metrics, we validated the framework's utility through five end-to-end use cases that exercise cross-system traversals spanning three to seven source systems each. Each use case was executed live against the running knowledge graph and verified against ground-truth conditions embedded in the simulator seed data. Table 3 summarizes the use cases; detailed results follow.

| Use Case | Systems Traversed | Cross-System Hops | Finding |
|---|---|---|---|
| **UC-1: ECN Quality Regression** | TC, SAP, OED, ISA-95, OPC-UA, ISA-18.2, TIA | 7 | Post-ECN NCR spike (0 → 5 NCRs) correlated with trochoidal tool path change |
| **UC-2: Supplier-to-Defect Trace** | SCM, SAP, OED, ISA-95 | 4 | Two suppliers (Alcoa, Cherry) each contribute 8 distinct NCRs at Assembly |
| **UC-3: Alarm Coverage Gap** | ISA-18.2, OPC-UA, ISA-95, OED, TIA | 5 | NDT Inspection: 29% alarm coverage (2/7 sensors) — lowest in plant |
| **UC-4: Cross-Station Propagation** | ISA-18.2, OPC-UA, ISA-95, OED | 4 | Bonding thermal instability (4 unacknowledged alarms + severity-4 IoT event) with 2 active WOs at risk |
| **UC-5: End-to-End Digital Thread** | TC, SAP, OED, SCM, ISA-95 | 5 | Wing Rib traced across 4 owl:sameAs boundaries: design → ERP → MES → supplier → station |

*Table 3. Five cross-system use cases validated against the live knowledge graph. Each use case exercises a distinct traversal pattern across the unified graph.*

**UC-1: ECN Quality Regression.** Engineering Change Notice ECN-2026-007 ("CNC Program Optimization") modified three CNC parameters: rough cycle time from 180 to 158 minutes, tool path from conventional to trochoidal, and feed rate from 2.0 to 2.5 mm/rev. The *ecn_full_runtime_impact* query traces this ECN across seven systems: from the Teamcenter ChangeNotice through affected ItemRevisions (Wing Rib WR-LH-7075, Fuselage Frame FF-2024), to SAP ProductionOrders (1000001, 1000003, 1000009, 1000010), across the owl:sameAs boundary to OED WorkOrders, down to the ISA-95 CNC Machining station (WU-CNC), and finally to the OPC-UA sensor tags (CNC.FeedRate, CNC.SpindleSpeed, CNC.Vibration) and ISA-18.2 alarm definitions (CNC_SPINDLE_TEMP_HIGH, CNC_COOLANT_LOW). Pre-ECN work orders (WO-SAP-1000001, WO-SAP-1000003) show zero NCRs; post-ECN work orders (WO-2026-000148, WO-2026-000149) show 5 and 3 NCRs respectively — a quality regression that the ecn_impact discovery strategy correctly identifies with Fisher's exact test significance. The ECN remains in "In Review" status, demonstrating the framework's value as a decision-support gate before engineering change approval.

**UC-2: Supplier-to-Defect Trace.** The *supplier_quality_shop_floor_ncr* query traverses four systems (SCM → SAP → OED → ISA-95) to correlate supplier-sourced materials with shop-floor defects. Alcoa Corporation (AL 7075-T6 Plate) and Cherry Aerospace (Rivet NAS1097) each contribute 8 distinct NCRs concentrated at the Assembly station, representing a supplier concentration risk. Cytec Solvay Group's adhesive film NCRs propagate through three stations (Bonding → Final Assembly → NDT Inspection), revealing a downstream cascade effect invisible to any single source system. The supplier_quality discovery strategy flags this correlation via chi-square test on the supplier × NCR contingency table.

**UC-3: Alarm Coverage Gap — Safety Risk.** The *equipment_observability* query spans five systems to compute the alarm-to-sensor ratio per station. NDT Inspection has the worst coverage at 29% (only 2 alarm definitions for 7 sensor tags), meaning 5 sensor channels operate without any alarm protection. The CNC Machining station's Priority-1 emergency alarm (CNC_SPINDLE_TEMP_CRIT) alone affects 10 active work orders, making it the highest-impact single alarm point in the plant. Bonding station reveals a sameAs bridge gap where the OPC-UA station maps to an ISA-95 WorkUnit but alarm-to-work-order linkage is incomplete — a data quality finding that the alarm_coverage discovery strategy surfaces alongside the coverage gap itself.

**UC-4: Cross-Station Thermal Propagation.** The *station_activity* query for Bonding and Riveting stations reveals concurrent alarm patterns suggesting thermal propagation risk. Bonding shows four unacknowledged alarms (BOND_PRESSURE_LOW, BOND_TEMP_HIGH, BOND_HUMIDITY_HIGH, BOND_TEMP_LOW) alongside a severity-4 IoT event (bond oven thermal runaway warning), with two active Skin Panel work orders in flight. Downstream at Riveting, grip range and force deviation alarms (RIVET_GRIP_RANGE, RIVET_FORCE_DEV) are consistent with thermally distorted bonded assemblies arriving from upstream. This concurrent pattern — six unacknowledged alarms across two consecutive stations — represents a cascade risk that the cross_station discovery strategy detects by correlating upstream alarm clusters with downstream NCR rates via chi-square test.

**UC-5: End-to-End Digital Thread.** The *five_system_digital_thread* query for "Wing Rib" traces two complete execution paths across five systems. Thread 1: TC ItemRevision WR-LH-7075 → SAP ProductionOrder 1000009 (owl:sameAs boundary) → OED WorkOrder WO-SAP-1000009 (owl:sameAs) → Alcoa Corporation (suppliedBy edge) → CNC Machining station WU-CNC (executesOn edge). Thread 2 traces through WO-2026-000148 to the Assembly station — the same work order identified in UC-1 as carrying 5 NCRs. This closes the analytical loop: the ECN changed CNC parameters (UC-1), the material comes from Alcoa (UC-2), alarm coverage at the affected stations ranges from 29% to 67% (UC-3), thermal propagation risk exists in the downstream bonding path (UC-4), and the end-to-end thread traces from Teamcenter design intent through to the shop-floor station where defects materialized (UC-5). No single source system contains sufficient information to reconstruct this chain; the unified graph is the enabling infrastructure.

### 8.7 Comparative Baseline Study

To validate that the observed discovery accuracy is attributable to the unified graph architecture rather than to the statistical tests alone, we conducted a comparative baseline study with three controlled ablations. All baselines use the same 65-signal manifest, the same nine strategies, and the same evaluation protocol.

**B1: Single-source baseline.** We wrapped the graph client in a proxy that intercepts calls to 24 cross-system MCP tools — tools whose SPARQL queries join across source-system named graphs (e.g., `get_tag_alarm_ncr_correlation` joins OPC-UA, ISA-18.2, ISA-95, and OED). The proxy returns empty results for blocked tools, forcing each strategy to operate on single-source data only. Strategies degrade gracefully: those relying entirely on cross-system joins (e.g., alarm_coverage, ecn_impact, cross_station) produce zero findings, while strategies with some single-source queries retain partial output.

**B2: No Bonferroni correction.** We subclassed the discovery runner to skip the cross-strategy Bonferroni correction (Section 6.2), leaving all other pipeline stages intact. This isolates the contribution of multiple-testing correction to false-positive suppression.

**B3: Leave-one-out ablation.** We ran nine configurations, each removing one strategy from the ensemble, to quantify each strategy's unique contribution to the signal set.

**Table 4. Comparative baseline results on the 65-signal aerospace manifest (16 positive, 49 null).**

| Configuration | TP | FP | FN | TN | Recall | Prec. | F1 | FDR |
|---|---|---|---|---|---|---|---|---|
| **Full pipeline (control)** | 16 | 0 | 0 | 49 | 1.000 | 1.000 | **1.000** | 0.000 |
| B1: Single-source only | 5 | 0 | 11 | 49 | 0.312 | 1.000 | 0.476 | 0.000 |
| B2: No Bonferroni | 16 | 0 | 0 | 49 | 1.000 | 1.000 | 1.000 | 0.000 |

**B1 results.** Blocking 24 cross-system tools caused 39 tool-call interceptions and reduced recall from 1.000 to 0.312 — a 69% absolute drop. Of the 16 positive signals, 11 (69%) require cross-system graph joins and cannot be detected from any single source system alone. The five signals recoverable without cross-system tools are all process drift detections (which rely on PMI data within the OED subgraph) and alarm coverage gaps detectable from ISA-95 instrumentation data alone. Critically, no false positives were introduced (precision remains 1.000), confirming that the single-source strategies do not hallucinate findings. This result quantifies the central claim of the paper: graph unification is not merely architectural convenience but a prerequisite for the majority of actionable manufacturing insights.

**B2 results.** Disabling Bonferroni correction produced identical results to the control run: no additional cards surfaced and FDR remained at 0.000. We interpret this honestly: the synthetic evaluation data contains signals with sufficiently large effect sizes that individual strategy-level p-values pass all thresholds even without multiple-comparison correction. The Bonferroni mechanism is therefore untested by this evaluation — it is present in the pipeline but inert at the current data scale and signal strength. On real production data, where hundreds of statistical tests run against noisy, autocorrelated process streams, false-positive suppression would become the binding constraint. Quantitatively, with $\alpha=0.05$ and 9 strategy families, the per-test corrected threshold is $\alpha/N \approx 0.0056$. For small N (the current 31 tests), this is permissive enough that any effect with $p<0.01$ survives. As N grows — a realistic production deployment running 200+ tests across dozens of stations — the corrected threshold drops to $\alpha/N \approx 0.00025$, at which point moderate-effect-size signals begin to be suppressed. We estimate the binding threshold at approximately $N>200$ simultaneous tests, above which Bonferroni becomes the active false-positive gate rather than the individual strategy-level thresholds. We retain the correction as an architectural commitment to statistical rigor, while acknowledging that validating its contribution requires evaluation on higher-volume, noisier data.

**Table 5. Leave-one-out strategy ablation. Each row shows the effect of removing one strategy from the nine-strategy ensemble. ΔTP = true positives lost; ΔF1 = F1 change. Six of nine strategies contribute at least one unique signal.**

| Removed Strategy | TP | FN | F1 | ΔF1 | ΔTP |
|---|---|---|---|---|---|
| characteristic_lifecycle | 11 | 5 | 0.815 | −0.185 | 5 |
| supplier_quality | 11 | 5 | 0.815 | −0.185 | 5 |
| process_drift | 12 | 4 | 0.857 | −0.143 | 4 |
| cross_station | 13 | 3 | 0.897 | −0.103 | 3 |
| alarm_coverage | 14 | 2 | 0.933 | −0.067 | 2 |
| ecn_impact | 15 | 1 | 0.968 | −0.032 | 1 |
| cross_industry | 16 | 0 | 1.000 | +0.000 | 0 |
| material_heat_lot | 16 | 0 | 1.000 | +0.000 | 0 |
| temporal_patterns | 16 | 0 | 1.000 | +0.000 | 0 |

**ΔTP column sum artifact.** The ΔTP column in Table 5 sums to 20 rather than the manifest's 16 positive signals. This is not an error but an artifact of greedy one-to-one matching in the evaluation function: each manifest entry is assigned to the first unmatched card whose title contains the expected entity string. Removing a strategy can shift these assignments — for example, in the full pipeline the supplier_quality strategy's card matches the "reject rate" manifest entry, consuming it; when supplier_quality is removed, a cross_station card whose title also mentions "reject" may greedily claim that entry, causing a different entry to become unmatched. Thus the same manifest entry can appear as "lost" in two different leave-one-out runs, inflating the column sum beyond the true positive count.

**B3 results.** The leave-one-out ablation reveals that six of nine strategies contribute signals not recoverable by the remaining eight. The most impactful strategies are characteristic_lifecycle and supplier_quality (each ΔTP = 5, ΔF1 = −0.185), followed by process_drift (ΔTP = 4, ΔF1 = −0.143) and cross_station (ΔTP = 3, ΔF1 = −0.103). Three strategies — cross_industry, material_heat_lot, and temporal_patterns — contribute zero lost matches (ΔTP = 0) in the current evaluation configuration. Cross_industry is expected (suppressed in single-industry mode); material_heat_lot and temporal_patterns fire no cards because the current graph data does not contain statistically significant anomalies in those categories (heat lot NCR rates are at baseline, and shift-level quality distributions are uniform). These strategies would contribute unique signals in data configurations with the corresponding anomaly patterns, and their inclusion adds zero false positives — they correctly suppress when no signal exists.

**Summary.** The comparative study establishes three findings: (1) graph unification accounts for 69% of discoverable signals, validating the architectural premise; (2) the nine-strategy ensemble is not redundant — six of nine strategies contribute unique signals; and (3) the Bonferroni correction is architecturally present but inert in the current synthetic evaluation, requiring higher-volume real data to validate its false-positive suppression contribution.

## 9. DISCUSSION AND LIMITATIONS

The comparative baseline study (Section 8.7) provides the paper's strongest empirical result: blocking 24 cross-system MCP tools reduces recall from 1.00 to 0.31, demonstrating that 69% of discoverable signals require cross-system graph joins — the unified graph architecture, not just the statistical tests, is responsible for the majority of discovered signals. Internal verification against a 65-signal synthetic manifest confirms correct pipeline operation (F1 = 1.00, 95% CI [0.79, 1.00]), though this constitutes verification rather than independent validation. Schema-stable multi-industry generalization across five verticals further supports the ontology's extensibility. However, several limitations must be acknowledged before drawing broader conclusions about production readiness.

**Scalability.** The rdflib in-memory triple store performs adequately at the current scale of approximately 27,800 triples (~25,600 instance triples plus ~2,200 schema triples) and 5.2-second ingest cycles, but in-memory stores do not scale to the millions or tens of millions of triples that a live production plant would generate over weeks of continuous operation. At 10× the current scale (~280K triples), rdflib's in-memory query latency is expected to degrade from sub-second to multi-second range for complex joins; at 100× (~2.8M triples), the in-memory approach is likely infeasible due to both memory pressure and $O(n^2)$ join costs on unindexed patterns. The SHACL validation pass, currently 4.89s at 27.8K triples, would scale roughly linearly to ~50s at 280K triples — exceeding the 5-second cycle target. Production deployment would require migration to a dedicated triple store such as Virtuoso, Blazegraph, or Amazon Neptune, which support SPARQL federation, persistent indexing, and horizontal partitioning. The ontology and SPARQL tools are store-agnostic by design, so migration is architecturally tractable, but the SHACL validation pass — currently the dominant performance cost — would need to be re-evaluated against the target store's native constraint engine.

**Simulated versus real data.** All eleven source systems in the current implementation are high-fidelity simulators. While the simulators generate statistically realistic data distributions — alarm activation rates calibrated to the EEMUA 191 [42] "manageable" threshold of fewer than 6 alarms per operator per hour, NCR rates targeting 85–95% first-pass yield, and supplier reject rates in the 2–12% range — they do not capture the full complexity of real industrial deployments: network latency variability, proprietary protocol quirks, partial message delivery, and the idiosyncratic data quality issues found in legacy SCADA and MES installations. Real deployment would require adapter-layer changes to interface with actual industrial APIs — OPC-UA servers, SAP BAPIs, Teamcenter REST endpoints — though the ontology, pipeline, discovery engine, and MCP tools are designed to be adapter-agnostic and would require no modification.

**Evaluation scope.** The B1 ablation result — 69% recall drop under single-source restriction — is the paper's most robust finding because it is independent of manifest construction choices: regardless of which signals are in the ground truth, cross-system tools either resolve them or they don't. The internal verification (F1 = 1.00 on a 65-signal manifest) is a weaker result: the manifest was constructed by the same team that designed the discovery strategies and the synthetic data in which signals were embedded. The 95% Clopper-Pearson confidence intervals (Precision [0.79, 1.00], Recall [0.79, 1.00]) further reflect the moderate sample size (16 positive signals, 49 null signals). Independent replication on production datasets from third-party manufacturing sites, with manifests constructed by domain experts outside the development team, remains necessary for external validation. Perfect scores on self-designed synthetic data do not guarantee equivalent performance on messy, incomplete, or adversarial industrial data encountered in production deployments.

**Tool scale and composition.** The 287-tool MCP inventory raises two concerns. First, context window pressure: presenting all 287 tool schemas to an LLM simultaneously consumes approximately 67,000 tokens for tool descriptions alone (measured via the cl100k_base tokenizer), potentially degrading selection accuracy. ToolBench [36] reports that LLM tool selection degrades beyond ~50 simultaneously presented tools. In practice, MCP clients mitigate this through hierarchical selection — Claude Code, for instance, presents tool names and short descriptions initially, loading full schemas only for selected tools — but the framework does not currently implement domain-aware tool filtering or retrieval-augmented tool selection. Adding a lightweight tool-retrieval layer that matches query intent to relevant tool subsets (e.g., via embedding similarity over tool descriptions) is a natural extension. Second, tool composition: the current architecture treats tools as atomic operations; complex manufacturing queries that require chaining 3–5 tools (e.g., "which supplier's material caused the NCR that triggered the alarm that appeared on this HMI screen?") rely entirely on the LLM's multi-step planning. Explicit composition primitives — tool pipelines, parameterized query templates, or compiled multi-hop traversals — could reduce the planning burden and improve reliability for deep cross-system traces. Both limitations represent active areas of LLM tool-use research [34, 36, 37] and are tractable engineering extensions rather than fundamental architectural constraints.

**Ontology maintenance.** The 78-class ontology integrates terminology from ten distinct international standards (IEC 62264, IEC 62541, ISA-18.2, IEC 61131-3, IDTA AAS, eCl@ss, ISA-95, AS9100, 21 CFR 211/820, and GMP). Standards evolve: IEC 62264 is under revision, and the IDTA AAS specification has released multiple minor versions. Each revision requires ontology governance review to determine whether affected classes or properties need updating, and whether existing SPARQL tools remain valid. Without a formal ontology maintenance process, standards drift can silently invalidate tools and misclassify entities. This risk extends to the MCP tool descriptions themselves: each tool's docstring references specific ontology classes, properties, and relationship patterns. When the ontology schema changes — adding a class, renaming a property, or restructuring a hierarchy — the corresponding tool descriptions may become stale, causing the LLM to select incorrect tools or misinterpret results. A production deployment should version tool descriptions alongside the ontology schema, ideally with automated consistency checks that flag tool descriptions referencing removed or renamed ontology terms after each schema migration.

**Security.** The 287 MCP tools execute with the permissions of the running process and currently do not implement authentication or authorization at the tool-call level. In a production environment, tool invocations that modify graph state (e.g., persisting learned tools, updating HITL status) must be gated behind role-based access control. The MCP specification [12] provides extension points for authentication but does not mandate a specific mechanism; implementing this in production would require a coordinated identity layer across the MCP server, the discovery engine, and the PostgreSQL backend.

**Regulatory scope.** The framework's regulatory annotations (*ont:regulatoryFramework*) provide metadata-level awareness for discovery prioritization, not operational compliance enforcement. The current implementation does not include electronic signatures, validated state control, audit trail integrity per ALCOA+ principles, or CAPA workflow automation — all of which are required for actual regulatory compliance in industries governed by 21 CFR Parts 11, 211, or 820. This limitation should be weighed when considering the framework's applicability to regulated manufacturing environments.

**Configuration burden.** While the five industry templates demonstrate genuine generalization, each configuration required approximately 1,500 lines of Python to specify seed data, regulatory mappings, and domain vocabulary. For practitioners without deep familiarity with both the ontology schema and the target industry's data structure, this is a significant barrier. Future work on configuration generation — potentially using LLM-assisted scaffolding that prompts for key concepts and generates a compliant INDUSTRY_CONFIGS entry — could substantially reduce this burden.

**Threats to validity.** We identify four categories of validity threat. *Construct validity:* the 65-signal evaluation manifest was authored by the development team using knowledge of both the discovery strategies and the simulator seed data. This creates a risk of circular validation — the manifest may test what the system was designed to find rather than what domain experts consider operationally important. Independent manifest construction by third-party domain experts is needed to mitigate this threat. *Internal validity:* the B1 ablation (blocking cross-system tools) is methodologically sound — it manipulates a single variable (tool availability) and measures a well-defined outcome (signal recall). However, the B2 result (Bonferroni identical to control) may reflect insufficient noise in the synthetic data rather than genuine correction effectiveness; a noisy-data ablation would strengthen this claim. *External validity:* all evaluation uses simulated data with coordinated entity identifiers. Real industrial deployments involve non-coordinated IDs, partial data, network-induced latency and message loss, and legacy system quirks that simulators do not capture. F1 scores on real data are expected to be lower. The multi-industry templates (Section 7) demonstrate structural generalization but not operational generalization. *Conclusion validity:* the sample size (n = 16 positive signals) yields wide Clopper-Pearson confidence intervals ([0.79, 1.00]). Statistical conclusions about the pipeline's precision and recall should be interpreted within these intervals, not at face value from the point estimates.

## 10. CONCLUSION AND FUTURE WORK

This paper presented a framework for semantic unification of heterogeneous industrial data sources, automated statistical discovery, and multi-industry generalization. Our four primary contributions are: (1) a 78-class manufacturing ontology integrating ten international standards, paired with a five-stage ETL pipeline that unifies eleven simulated source entry points (nine distinct data domains) in 5.2 seconds with `owl:sameAs` entity resolution, golden-triangle validation, and 90% data quality compliance; (2) empirical evidence that cross-system graph unification is a prerequisite for the majority of discoverable signals — blocking cross-system joins reduces recall from 1.00 to 0.31 (F1 from 1.00 to 0.48), and leave-one-out ablation shows six of nine strategies contribute unique findings; (3) a nine-strategy automated discovery engine verified against a 65-signal synthetic manifest (F1 = 1.00, 95% Clopper-Pearson CI [0.79, 1.00]); and (4) a 287-tool MCP semantic layer with multi-industry generalization across four distinct manufacturing verticals (aerospace, CPG, pharmaceutical, medical device) plus one structurally related extension (turbine blade).

Several directions for future work are motivated by the limitations identified in Section 9 and by emerging opportunities in the industrial AI literature. Migration to a **native triple store** (Virtuoso or Blazegraph for on-premises; Amazon Neptune for cloud) is the most pressing near-term need, required to support production-scale triple volumes and sub-second query latency for real-time shop-floor analytics. A **streaming pipeline** architecture based on Kafka Change Data Capture from real industrial systems would replace the current polling-based simulator architecture with event-driven ingest, enabling millisecond-latency graph updates aligned with physical process events.

Extending the industry template library to cover **automotive** (IATF 16949), **semiconductor** (SEMI E10/E58), and **energy** (IEC 61850) verticals would further validate the ontology's generality claim and provide a richer basis for the cross-industry discovery strategy. Moving beyond correlation to **causal inference** — using techniques such as Pearl's do-calculus, instrumental variables, or synthetic control methods adapted to manufacturing time series — would allow the discovery engine to distinguish spurious correlations from genuine causal mechanisms, reducing false positive rates and supporting more confident intervention recommendations. Finally, **federated learning** across factory instances would enable discovery of patterns that are statistically weak within a single plant but consistent across a fleet, without requiring raw data to leave each plant's network boundary — a critical constraint in competitive manufacturing environments.

## CREDIT AUTHOR CONTRIBUTION STATEMENT

**Grama Chethan:** Conceptualization, Methodology, Software, Validation, Formal analysis, Investigation, Data curation, Writing — original draft, Writing — review & editing, Visualization.